\documentclass[10pt,twocolumn,letterpaper]{article}

\usepackage{xcolor}
\usepackage{soul}
\usepackage[pagenumbers]{cvpr} 

\newcommand{\todo}[1]{{\color{red}#1}}

\definecolor{junglegreen}{rgb}{0.113, 0.639, 0.5}

\definecolor{magenta}{rgb}{0.8, 0.2 ,1}

\usepackage{microtype}

\definecolor{cvprblue}{rgb}{0.21,0.49,0.74}
\usepackage[pagebackref,breaklinks,colorlinks,allcolors=cvprblue]{hyperref}

\def\paperID{9831} 
\def\confName{CVPR}
\def\confYear{2026}

\title{RelightAnyone: A Generalized Relightable 3D Gaussian Head Model}

\author{Yingyan Xu$^{1, 2}$
	\hspace{7mm}
	Pramod Rao$^{3}$
	\hspace{7mm}
	Sebastian Weiss$^{2}$
    \hspace{7mm}
    Gaspard Zoss$^{2}$ \\
	Markus Gross$^{1, 2}$
	\hspace{7mm}
	Christian Theobalt$^{3}$
    \hspace{7mm}
    Marc Habermann$^{3}$
    \hspace{7mm}
	Derek Bradley$^{2}$
	\\
	$^{1}$ETH Z\"urich
	\hspace{7mm}
	$^{2}$DisneyResearch\textbar Studios 
    \hspace{7mm}
    $^{3}$ Max Planck Institute for Informatics, SIC \\
}

\begin{document}

\twocolumn[{%
	\maketitle
	\begin{center}
		\centering
		\vspace{-5mm}
          \includegraphics[width=\linewidth]{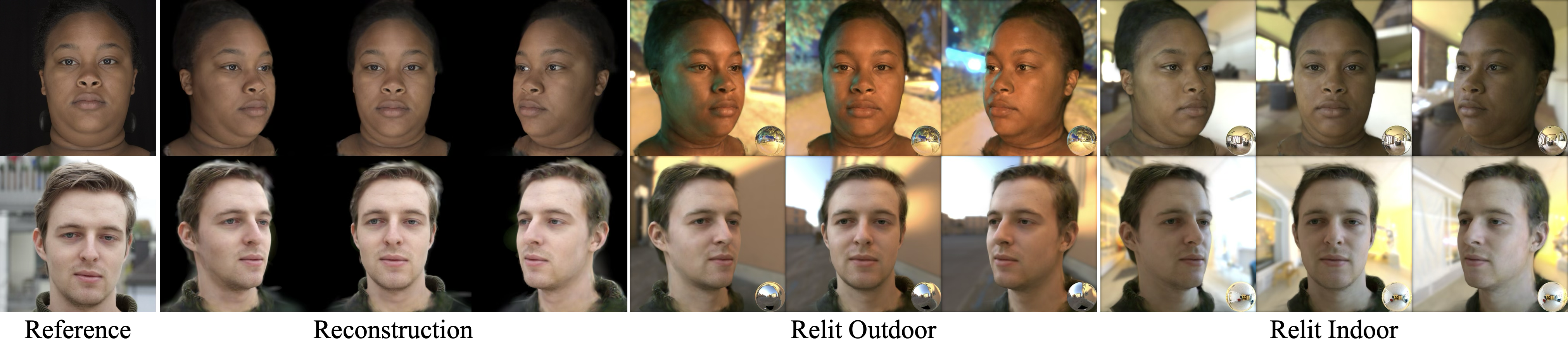}
		\captionof{figure}{We present RelightAnyone, a method for reconstructing and relighting head avatars from multi-view (top row) or single images in-the-wild (bottom row), trained on a limited dataset of multi-view OLAT faces and larger datasets of mulit-view flat-lit faces.}
		\label{fig:teaser}
	\end{center}%
}]

\newcommand{\figref}[1]{Fig.~\ref{#1}}
\newcommand{\tabref}[1]{Table~\ref{#1}}
\newcommand{\eqnref}[1]{Eq.~\ref{#1}}
\newcommand{\secref}[1]{Section~\ref{#1}}
\newcommand{\appref}[1]{Appendix~\ref{#1}}


\newcommand{\shortcite}[1]{\cite{#1}}


\definecolor{dbcolor}{RGB}{50,10,210}
\newcommand\db[1] {{\textcolor{dbcolor}{\em\textbf{DB}: #1}}}

\definecolor{swcolor}{RGB}{210,10,210}
\newcommand\sw[1] {{\textcolor{swcolor}{\em\textbf{SW}: #1}}}

\definecolor{yxcolor}{RGB}{10,210,50}
\newcommand\yx[1] {{\textcolor{yxcolor}{\em\textbf{YX}: #1}}}

\definecolor{gzcolor}{RGB}{50,210,210}
\newcommand\gz[1] {{\textcolor{gzcolor}{\em\textbf{GZ}: #1}}}

\definecolor{delcolor}{RGB}{210,0,0}
\definecolor{addcolor}{RGB}{0,0,0}
\newcommand\del[1] {{\textcolor{delcolor}{#1}}}
\newcommand\add[1] {{\textcolor{addcolor}{#1}}}

\renewcommand\todo[1] {{\textcolor{red}{\em\textbf{TODO}: #1}}}
\newcommand\camready[1] {{\textcolor{blue}{#1}}}

\makeatletter
\newcommand\footnoteref[1]{\protected@xdef\@thefnmark{\ref{#1}}\@footnotemark}
\makeatother

\renewcommand{\thefootnote}{\fnsymbol{footnote}}

\clubpenalty=10000
\widowpenalty=10000
\displaywidowpenalty=10000

%
%
\begin{abstract}
3D Gaussian Splatting (3DGS) has become a standard approach to reconstruct and render photorealistic 3D head avatars.
A major challenge is to relight the avatars to match any scene illumination.
For high quality relighting, existing methods require subjects to be captured under complex time-multiplexed illumination, such as one-light-at-a-time (OLAT).
We propose a new generalized relightable 3D Gaussian head model that can relight any subject observed in a single- or multi-view images without requiring OLAT data for that subject.
Our core idea is to learn a mapping from flat-lit 3DGS avatars to corresponding relightable Gaussian parameters for that avatar.
Our model consists of two stages: 
a first stage that models flat-lit 3DGS avatars without OLAT lighting, and a second stage that learns the mapping to physically-based reflectance parameters for high-quality relighting.
This two-stage design allows us to train the first stage across diverse existing multi-view datasets without OLAT lighting ensuring cross-subject generalization, where we learn a dataset-specific lighting code for self-supervised lighting alignment.
Subsequently, the second stage can be trained on a significantly smaller dataset of subjects captured under OLAT illumination.
Together, this allows our method to generalize well and relight any subject from the first stage as if we had captured them under OLAT lighting. 
Furthermore, we can fit our model to unseen subjects from as little as a single image, allowing several applications in novel view synthesis and relighting for digital avatars.
%
\end{abstract}
%
%
%
%
\section{Introduction}
\label{sec:intro}
%
%
For decades, researchers have strived to create photorealistic digital head avatars from images of real people.
With the recent introduction of 3D Gaussian Splatting (3DGS) \cite{kerbl20233d}, the problem has become easier.
3DGS allows efficient scene reconstruction from multiple views, and then real-time rendering from any novel angle.
As such, by now there are several methods~\cite{qian2024gaussianavatars,Xu2023gaussianheadavatar,Teotia2024gaussianheads,Aneja2025scaffoldavatar} for reconstructing and animating 3D avatars using 3DGS.
%
%
\par 
A main challenge that prevents the widespread use of 3DGS for deploying digital avatars across diverse virtual environments is the lack of disentanglement of human facial reflectance from scene illumination.
Such avatars require the ability to decompose the illumination into a subject-specific albedo, together with diffuse and specular reflection parameters, to enable efficient relighting, allowing us to place the avatar into any virtual environment.
The seminal work of Debevec~\etal \cite{debevec2000acquiring} demonstrates that faces captured under an OLAT lighting condition within a light stage enable high quality physically accurate face reflectance modeling that is fit for relighting.
Recent works like \emph{Relightable Gaussian Codec Avatars (RGCA)}~\cite{saito2024relightable} leverage OLAT-based captures and extend 3DGS with parameters separating surface material from scene illumination to achieve efficient 3D avatar relighting.
Specifically, a model is trained to predict the extended 3DGS parameters in the texture-space of a coarse template mesh.
The prediction includes learned radiance transfer functions (\eg diffuse color, specular roughness, normals, etc.) so that the avatar can be relit under desired environment lighting.
While results are impressive, the learned avatars are person-specific; each new subject requires another lightstage capture and model retraining, which is both labor- and compute-intensive.
%
%
\par 
Similar to URAvatar~\cite{li2024uravatar}, in this work we aim to generalize this approach to allow relighting of anyone, not captured under time-multiplexed illumination.
However, URAvatar heavily relies on capturing a large number of diverse subjects under OLAT lighting for training their prior model. 
Unfortunately, building such a large OLAT datasets is very expensive and time-consuming.
On the other hand, capturing subjects under fixed lighting is much easier and a significant number of diverse identities are already available in multi-view flat-lit human head datasets (\eg Ava-256~\cite{martinez2024codec}, Nersemble~\cite{kirschstein2023nersemble}). 
Our core idea is to leverage existing flat-lit datasets for identity generalization, together with a comparably smaller amount of public OLAT data~\cite{prao20253dpr} for relighting.
%
%
\par 
Specifically, we accomplish this by learning a mapping from fixed, flat-lit 3DGS avatars to the corresponding relightable RGCA parameters.
Our model consists of two stages.
In the first stage, we train a network to predict a 3DGS avatar under fixed lighting, conditioned on the subject identity.
This includes an MLP to predict the coarse mesh shape, and CNNs to predict Gaussian parameters in texture space.
We train this network on several different fixed-lighting datasets to ensure generalization to new identities. 
However, each dataset comes with the challenge that it has different illumination conditions and camera parameters.
Thus, learning a uniform neutral color space for relighting is hard. 
Therefore, we propose a learned dataset-specific lighting code, which we optimize via our self-supervised lighting alignment.
In a second stage, we then introduce a UNet to map from flat-lit 3DGS parameters to RGCA parameters for relighting.
Importantly, we show that this second network can be trained on a comparably smaller OLAT dataset while not compromising identity generalization. 

Once trained, our method allows several applications for generalized avatar reconstruction and relighting.
First, we can trivially relight any subject seen in the fixed-lighting datasets, leading to the interesting application of creating synthetic OLAT renders and expanding existing fixed-light datasets to OLAT datasets.
Second, we can fit our model to unseen subjects under unseen lighting conditions.
We demonstrate fitting and relighting subjects from as little as a single input image in the wild, yielding a powerful approach to build avatars that can be relit and rendered from any novel view. 
In summary, our main contributions are:
%
\begin{itemize}
    \item A relightable and generalizable Gaussian head model with a novel two-stage pipeline enabling unified training across diverse multi-view datasets, both flat-lit and with OLAT lighting. 
    \item A method for self-supervised lighting alignment across flat-lit multi-view head avatar datasets, through the introduction of a learnable dataset-specific lighting code.
    \item A relighting network for mapping Gaussian colors under full-on lights to relightable Gaussian parameters. 
    \item A fitting approach that allows our model to be fitted to multi-view images or a single portrait photo in the wild, yielding high quality relightable 3DGS head avatars.
\end{itemize}

%
%
\section{Related Work}
\label{sec:related}

We first outline relevant work on 2D relighting, which is typically constrained to the original camera view. We then review methods more closely related to ours that address 3D human face relighting. 

\paragraph{2D Relighting.} Portrait relighting has been a long-standing research topic, with early methods primarily relying on deep convolutional architectures. While initial works implicitly learned the relighting process in a black-box manner~\cite{sun2019single, meka2019deep}, later methods moved toward more explicit, physics-based designs, incorporating image intrinsics and reflectance models directly into their architectures~\cite{nestmeyer2020learning, wang2020single, pandey2021total, ji2022geometry, mei2023lightpainter, Kim_2024_CVPR}. To address the challenge of acquiring large-scale light stage data, alternative solutions were proposed, such as reducing hardware requirements~\cite{sengupta2021light} or using synthetic data~\cite{yeh2022learning}.

More recently, the paradigm of 2D relighting has shifted towards generative methods, leveraging the success of image~\cite{dhariwal2021diffusion, rombach2022high} and video~\cite{blattmann2023stable, blattmann2023align, guo2023animatediff, singer2022make} diffusion models. These new approaches have been applied to portrait and scene relighting~\cite{ponglertnapakorn2023difareli, kocsis2024lightit, poirier2024diffusion, zeng2024dilightnet, he2024diffrelight, Mei_2025_CVPR}, as well as related tasks such as illumination harmonization~\cite{ren2024relightful, zhang2025scaling} and general inverse rendering~\cite{DiffusionRenderer, he2025unirelight}. While harmonization methods like IC-Light~\cite{zhang2025scaling} excel at matching a foreground to a background, they generally lack mechanisms for fine-grained, explicit lighting control via HDRI maps. Furthermore, general-purpose inverse rendering methods, such as DiffusionRenderer~\cite{DiffusionRenderer}, often fail to model the complex and unique reflectance of human skin, resulting in unrealistic material properties when applied to portraits.

\paragraph{3D Face Relighting.} Beyond image-to-image translation, a significant body of work has focused on 3D relighting, which enables simultaneous relighting and novel-view synthesis. Traditional methods capture the precise skin geometry and reflectance of a subject under complex, calibrated hardware, which is then used to create a high-fidelity, relightable avatar for that specific person~\cite{debevec2000acquiring, weyrich2006analysis, ma2007rapid, ghosh2011multiview, fyffe2016near, guo2019relightables, riviere2020single, xu2022improved}. This high-quality capture data has also been used to train person-specific neural representations. Early neural methods focused on learning relightable textures on 3D meshes~\cite{Meka:2020, zhang2021neural, bi2021deep}, while more recent works have adopted volumetric representations based on Neural Radiance Fields~\cite{mildenhall2021nerf, muller2022instant, xu2023renerf, sarkar2023litnerf}, Mixtures of Volumetric Primitives~\cite{lombardi2021mixture, yang2023towards, xu2024artist}, or 3DGS~\cite{kerbl20233d, saito2024relightable, schmidt2025becominglit}. A state-of-the-art example in this domain is RGCA~\cite{saito2024relightable}, which achieves exceptionally high-quality results by introducing a learnable radiance transfer for 3D Gaussians, enabling real-time relighting with all-frequency reflections. 

To bypass the need for expensive light stages, many works have investigated more accessible, light-weight setups. These approaches vary in their hardware requirements, from desktop setups~\cite{lattas2022practical}, to setups using co-located light and cameras~\cite{azinovic2023high, han2024high}. Others use the sun as a dominant point source~\cite{wang2023sunstage}, or create avatars from simple monocular inputs~\cite{dib2021practical, bharadwaj2023flare, rainer2023neural, xu2025monocular, HRAvatar}. Although more accessible, these in-the-wild optimization techniques still struggle to match the relighting fidelity of avatars captured in a light stage.

This quality gap, combined with the inherently ill-posed nature of these in-the-wild captures, motivated the development of generalized methods that learn a strong generative prior from large-scale datasets. One common approach is to learn intrinsic skin properties (\eg, surface normals, albedo, roughness)~\cite{lattas2020avatarme, dib2021towards, lattas2021avatarme++, dib2022s2f2, lin2023single, han2025facial}. However, this approach is often limited to the skin and is difficult to unify across the entire head. Another popular approach is to leverage 3D-aware GANs, such as EG3D~\cite{chan2022efficient} as a prior to synthesize relightable faces~\cite{tan2022volux, jiang2023nerffacelighting, ranjan2023facelit, rao2024lite2relight, deng2024lumigan, mei2024holo, prao20253dpr, lv2025gsheadrelight}. However, since these methods are often trained on only 2D portrait collections, their learned 3D geometry is often incomplete and lacks texture on the back of the head, leading to unrealistic or artifact-filled novel-view synthesis for non-frontal poses. A different line of work achieves better 3D consistency by learning generalizable volumetric representations from multi-view datasets~\cite{sun2021nelf, rao2022vorf, yang2024vrmm, li2024uravatar}. Closest to ours, URAvatar~\cite{li2024uravatar} builds upon RGCA, but it requires a large-scale, difficult-to-acquire dataset with multi-view OLAT capture for every subject. In contrast, our specially designed two-stage pipeline bypasses this data-acquisition bottleneck. We leverage existing, more common multi-view flat-lit datasets~\cite{chandran2020semantic, kirschstein2023nersemble, martinez2024codec} supplemented by only a small OLAT dataset~\cite{prao20253dpr}. Furthermore, URAvatar requires an unwrapped albedo texture for identity conditioning, which is non-trivial to obtain, especially in in-the-wild scenarios, and unwrapping can fail when only a single image is available. 
\section{Method}
\label{sec:method}

\begin{figure*}[t]
  \centering
  \includegraphics[width=\linewidth]{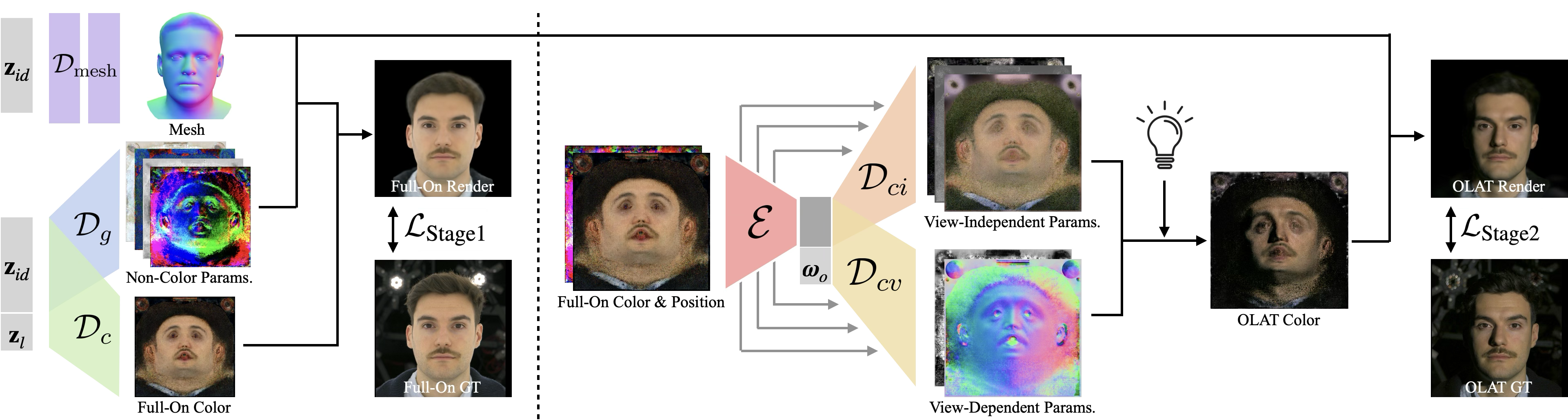}

   \caption{Pipeline of RelightAnyone: We first train a Stage 1 network that, given a learned subject identity code and a learned lighting code that distinguishes between different illuminations of different datasets, predicts a Gaussian avatar under full-on lighting, where the parameters of the Gaussian primitives are encoded in UV texture maps. Then, the Stage 2 relighting network maps the full-on Gaussian parameter textures from Stage 1 to RGCA parameters to allow for relighting of the avatar.}
   \label{fig:pipeline}
\end{figure*}

We first review RGCA~\cite{saito2024relightable} in~\secref{sec:rgca}. We then detail our proposed two-stage pipeline in~\secref{sec:two-stage} and, finally, describe fitting to unseen identities in~\secref{sec:fitting}.

\subsection{Preliminary: RGCA}
\label{sec:rgca}

A 3D Gaussian is parameterized by a translation vector $\mathbf{t}_k \in \mathbb
{R}^3$, a unit quaternion $\mathbf{q}_k \in \mathbb{R}^4$, scale factors $\mathbf{s}_k \in \mathbb{R}_+^3$, an opacity value $o_k \in [0,1]$, and a color $\mathbf{c}_k \in \mathbb{R}_+^3$. To make 3D Gaussians relightable, the color is parameterized to interact with incident lighting. More specifically, in RGCA~\cite{saito2024relightable}, the Gaussian color is computed as the sum of a diffuse color $\mathbf{c}_k^\text{diffuse}$ and a specular color $\mathbf{c}_k^\text{specular}$. The diffuse color is computed as
\begin{equation}
\label{eq:diffuse}
    \mathbf{c}_k^\text{diffuse} = \boldsymbol{\rho}_k \odot \sum^{(n+1)^2}_{i=1} \mathbf{L}_i \odot \mathbf{d}_k^i
\end{equation}
where $\boldsymbol{\rho}_k \in \mathbb{R}_+^3$ is the diffuse albedo. $\mathbf{L} = \{\mathbf{L}_i\}$ and $\mathbf{d}_k = \{\mathbf{d}_k^i\}$ are the $n$-th order spherical harmonics (SH) coefficients of the incident light and intrinsic radiance transfer function (where $\mathbf{d}_k^i \in \mathbb{R}^3$), respectively. The specular reflection is represented as a spherical Gaussian $G_s(\boldsymbol{\omega};\mathbf{a}_k, \sigma_k)$, defined by the lobe $\mathbf{a}$ and roughness $\sigma_k \in (0, 1)$. The final specular color from the viewing direction $\boldsymbol{\omega}_k^o$ is then calculated as:
\begin{align}
    \label{eq:spec}
    \mathbf{c}_k^\text{specular}(\boldsymbol{\omega}_k^o) &= v_k(\boldsymbol{\omega}_k^o) \int_{\mathbb{S}^2} \mathbf{L}(\mathbf{\omega})G_s(\boldsymbol{\omega};\mathbf{a}_k, \sigma_k)d\boldsymbol{\omega} \\
    \mathbf{a}_k &= 2(\boldsymbol{\omega}_k^o \cdot \mathbf{n}_k)\mathbf{n}_k - \boldsymbol{\omega}_k^o
\end{align}
where $v_k \in (0, 1)$ is a view-dependent visibility term, and $\mathbf{n}_k$ is a view-dependent specular normal. The integral in~\eqnref{eq:spec} can be efficiently evaluated for point sources represented with Dirac delta functions, or for prefiltered environment maps~\cite{kautz2000unified}. In summary, a non-relightable 3D Gaussian under full-on lighting $\mathbf{g}_k^f$, and its relightable counterpart $\mathbf{g}_k^r$ can be denoted as:
\begin{align}
\mathbf{g}_k^f &= \{\mathbf{t}_k, \mathbf{q}_k, \mathbf{s}_k, o_k, \mathbf{c}_k^f\} \\
\mathbf{g}_k^r &= \{\mathbf{t}_k, \mathbf{q}_k, \mathbf{s}_k, o_k, \boldsymbol{\rho}_k, \mathbf{d}_k, \sigma_k, v_k, \mathbf{n}_k\}
\end{align}

\subsection{RelightAnyone}
\label{sec:two-stage}

As shown in~\figref{fig:pipeline}, our two-stage pipeline first learns a multi-identity Gaussian avatar model under full-on lighting from different datasets (Stage 1). A subsequent Stage 2  relighting network then maps full-on Gaussian colors to relightable RGCA parameters. The network and training details are described below; please refer to the supplementary material for further implementation details. 

\paragraph{Stage 1: Multi-Identity Full-On Model.} 
Our Stage 1 architecture differs from RGCA by utilizing a learnable identity code $\mathbf{z}_\text{id} \in \mathbb{R}^{256}$ and a low-dimensional, dataset-specific lighting code $\mathbf{z}_l \in \mathbb{R}^4$. More specifically, it is composed of three decoders: a mesh decoder $\mathcal{D}_\text{mesh}$, a decoder $\mathcal{D}_g$ that outputs geometry-related Gaussian parameters, and a Gaussian color decoder $\mathcal{D}_c$. $\mathcal{D}_\text{mesh}$ is implemented as a multilayer perceptron. Both $\mathcal{D}_g$ and $\mathcal{D}_c$ are 2D convolutional neural networks that decode 3D Gaussians in a shared UV texture map of a coarse template mesh. Formally, we have:
\begin{align}
\mathbf{V} &= \mathcal{D}_\text{mesh}(\mathbf{z}_\text{id}) \\
\{\delta\mathbf{t}_k, \mathbf{q}_k, \mathbf{s}_k, o_k\}_{k=1}^M &= \mathcal{D}_g(\mathbf{z}_\text{id}) \\
\{\mathbf{c}_k^f\}_{k=1}^M &= \mathcal{D}_c(\mathbf{z}_\text{id}, \mathbf{z}_l)
\end{align}
Here, $\mathbf{V} \in \mathbb{R}^{5143 \times 3}$ represents the base mesh vertices. $\delta\mathbf{t}_k$ denotes the position offset of the 3D Gaussian \wrt the base mesh. The final Gaussian position $\mathbf{t}_k$ is computed as $\mathbf{t}_k = \hat{\mathbf{t}}_k + \delta\mathbf{t}_k$, where $\hat{\mathbf{t}}_k$ is derived by applying barycentric interpolation to the vertices using the corresponding UV coordinates. We use $M = 1024 \times 1024$ in all our experiments as the total number of Gaussians.

The lighting code $\mathbf{z}_l$ is first concatenated with $\mathbf{z}_\text{id}$ and fed into $\mathcal{D}_c$ to disentangle dataset-specific illumination properties. Although our datasets are generally evenly-lit, the exact lighting distributions still differ. 
$\mathbf{z}_l$ allows the network to separate this dataset-specific lighting impact from the canonical appearance of the subjects, leading to a ``cleaner" and more structured identity latent space (\figref{fig:setup-code}). This disentanglement also enables us to transfer lighting between datasets by simply swapping their $\mathbf{z}_l$ codes. This capability is essential because the Stage 2 relighting network is trained to map Gaussian colors to relightable parameters under one specific, full-on lighting condition corresponding to one dataset. Therefore, to use the relighting network, we must first generate the Gaussian colors $\mathbf{c}_k^f$ under that exact full-on condition for which it was trained.

We train our Stage 1 model with an L1 and an SSIM loss on the rendered images, a geometry reconstruction loss, and a scale regularization term $\mathcal{L}_\text{s}$, in a manner similar to the original RGCA. We introduce another term $\mathcal{L}_\text{t}$ that regularizes the Gaussian position offsets $\delta\mathbf{t}_k$ to be small:
\begin{equation} 
\mathcal{L}_\text{Stage1} = \lambda_\text{l1}\mathcal{L}_\text{l1} + \lambda_\text{ssim}\mathcal{L}_\text{ssim} + \lambda_\text{geo}\mathcal{L}_\text{geo} + \lambda_\text{s}\mathcal{L}_\text{s} + \lambda_\text{t}\mathcal{L}_\text{t}
\end{equation}

\paragraph{Stage 2: Relighting Network.}
Our Stage 2 model is a UNet that translates Gaussian colors under full-on illumination to relightable Gaussian parameters. This UNet features a shared encoder $\mathcal{E}$ and corresponding skip connections, which feed into two distinct decoder branches: a view-independent decoder $\mathcal{D}_{ci}$ and a view-dependent decoder $\mathcal{D}_{cv}$:
\begin{align}
\{\boldsymbol{\rho}_k, \mathbf{d}_k, \sigma_k\}_{k=1}^M &= \mathcal{D}_{ci}(\mathcal{E}(\mathbf{c}_k^f, \mathbf{t}_k))\\
\{v_k, \delta\mathbf{n}_k\}_{k=1}^M &= \mathcal{D}_{cv}(\mathcal{E}(\mathbf{c}_k^f, \mathbf{t}_k), \boldsymbol{\omega}_o)
\end{align}
Here, $\boldsymbol{\omega}_o$ is the viewing direction from the camera position to the center of the head mesh, and is concatenated to every pixel of the feature map at the network's bottleneck. The normal residual $\delta\mathbf{n}_k$ is added to the barycentric interpolated coarse mesh normal $\hat{\mathbf{n}}_k$ to obtain the final normal $\mathbf{n}_k$: $\mathbf{n}_k = (\hat{\mathbf{n}}_k + \delta\mathbf{n}_k) / ||\hat{\mathbf{n}}_k + \delta\mathbf{n}_k||$. 
Given any light conditions, these decoded RGCA parameters can then be used to compute the relit Gaussian colors by applying~\eqnref{eq:diffuse} and~\eqnref{eq:spec}. To enable the network to learn shape-dependent shading variations, we concatenate the 3D Gaussian positions $\mathbf{t}_k$ with full-on Gaussian colors, using this as three additional channels for the encoder's input. This is needed because, for example, if two subjects have identical base colors under full-on lighting but possess different facial geometries, they should look different under the same point light due to effects like self-shadowing. To learn a meaningful mapping, the relighting network must be trained on subjects captured under diverse and known lighting conditions (\eg, OLAT illuminations). However, calibrated multi-view \emph{and} lighting setups are expensive, and, thus, only a few public datasets exist with limited number of identities. We therefore train our Stage 2 model after the Stage 1 model is trained, rather than using an end-to-end approach. This sequential method allows training on both flat-lit and OLAT datasets while also keeping the pre-trained identity prior of Stage 1 intact. 

In RGCA, the albedo parameter $\boldsymbol{\rho}_k$ is not decoded by a network but is instead optimized jointly with the network parameters, starting from an initial mean texture. This approach is not feasible in our generalized case because we aim to predict the albedo for unseen subjects. However, allowing the network to predict the albedo in an unconstrained manner leads to a non-meaningful decomposition of the albedo and shading parameters. We therefore introduce two regularization terms to mitigate this issue. The first term, $\mathcal{L}_\rho$, is an L2 loss that regularizes the predicted albedo, encouraging it to stay close to the mean texture computed under full-on lighting. The second term, $\mathcal{L}_\text{mono}$, encourages the diffuse SH coefficients $\mathbf{d}_k$ to stay close to monochromatic:
\begin{equation}
\mathcal{L}_\text{mono} = \frac{1}{3}\sum_{i \in \{r, g, b\}}(\mathbf{d}_{ki} - \frac{\mathbf{d}_{kr} + \mathbf{d}_{kg} + \mathbf{d}_{kb}}{3})^2
\end{equation}
where $\mathbf{d}_{kr}$, $\mathbf{d}_{kg}$, and $\mathbf{d}_{kb}$ represent the RGB channels. The final loss function for Stage 2 training is then defined as:
\begin{equation}
\begin{split}
\mathcal{L}_\text{Stage2} = &\lambda_\text{l1}\mathcal{L}_\text{l1} + \lambda_\text{ssim}\mathcal{L}_\text{ssim} + \lambda_{c\_}\mathcal{L}_{c\_} \\
& + \lambda_n\mathcal{L}_n + \lambda_\rho\mathcal{L}_\rho + \lambda_\text{mono}\mathcal{L}_\text{mono}
\end{split}
\label{eqn:Lstage2}
\end{equation}
where $\mathcal{L}_{c\_}$~\cite{saito2024relightable} penalizes negative colors in the diffuse term as SH can yield negative values and $\mathcal{L}_n$ is an L2 loss regularizing the normal residual $\delta\mathbf{n}_k$ to be small.

\subsection{Model Fitting} 
\label{sec:fitting}
Our model can be fitted to unseen identities from single image or multi-view images using a two-step optimization process, similar to previous avatar personalization works~\cite{buhler2023preface, wang2022morf, yang2024vrmm, li2024uravatar}: first an inversion step to find an optimal identity code and scene lighting while keeping the networks frozen, followed by a finetuning step that updates the network parameters.

During the fitting process, the model is always executed as a full pipeline, combining Stage 1 and Stage 2, with the lighting code $\mathbf{z}_l$ set to the value associated with the dataset containing the OLAT data used to train Stage 2. The final image is rendered from the relightable Gaussians predicted by Stage 2, rather than the intermediate full-on Gaussians from Stage 1. This is crucial because the scene lighting is unknown and must be optimized as part of the fitting.

\paragraph{Inversion.} In the inversion step, we optimize the identity code and the scene lighting. We initialize $\mathbf{z}_\text{id}$ as the mean of the learned training subject codes. The lighting is parameterized as the same set of fixed-position point lights used in training. We optimize an RGB-intensity for each point light. The loss for this step combines image and shape reconstruction losses, as well as a L2 regularizer on $\mathbf{z}_\text{id}$:
\begin{equation}
    \mathcal{L}_\text{fit}^1 = \lambda_\text{l1}\mathcal{L}_\text{l1} + \lambda_\text{ssim}\mathcal{L}_\text{ssim} + \lambda_\text{geo}\mathcal{L}_\text{geo} + \lambda_\text{id} ||\mathbf{z}_\text{id}||^2
\end{equation}

\paragraph{Finetuning.} In the finetuning step, we further refine the Stage 1 network weights to capture person-specific details. We keep the Stage 2 network frozen to preserve the learned relighting prior. To prevent overfitting and maintain a plausible avatar structure, we incorporate an additional locality regularization loss $\mathcal{L}_\text{lr}$~\cite{yang2024vrmm}, a technique proven effective for prior preservation:
\begin{equation}
    \mathcal{L}_\text{fit}^2 = \mathcal{L}_\text{fit}^1 + \lambda_\text{lr}\mathcal{L}_\text{lr}
\end{equation}
\section{Experiments}
\label{sec:exp}

\begin{figure*}[t]
  \centering
  \includegraphics[width=\linewidth]{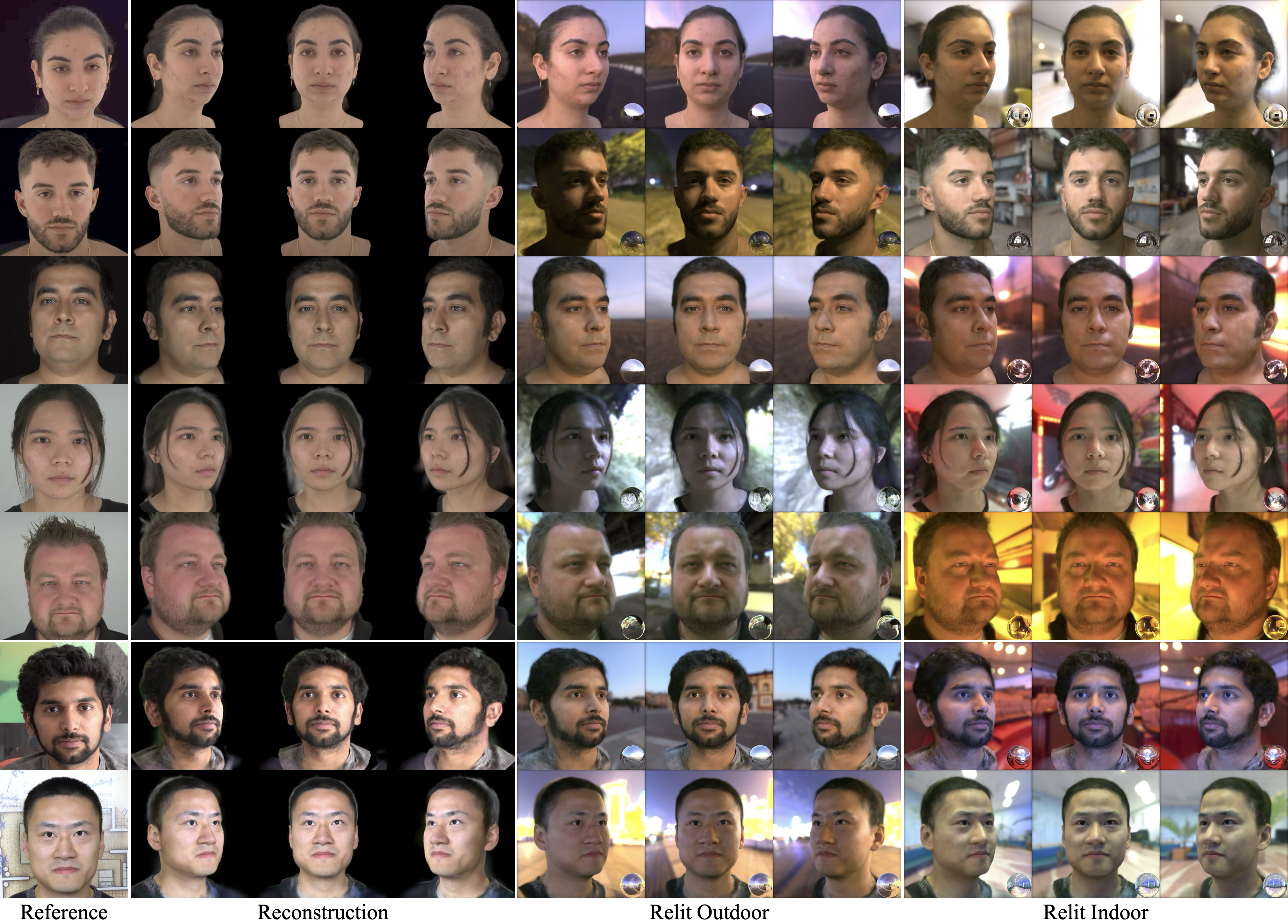}
   \caption{Reconstruction and relighting examples from multi-view (rows 1-5) and single images in the wild (rows 6-7). From top to bottom, the first two subjects are from D2; subject 3 is from D3; subject 4 and 5 are from D4, and subject 6 and 7 are self-captured in-the-wild images. Our method can accurately reconstruct a detailed and multi-view stable Gaussian avatar in Stage 1 (column ``Reconstruction'') that can be relit with arbitrary environment maps using Stage 2, even under harsh outdoor illumination. Images best viewed zoomed-in.}
   \label{fig:hero}
\end{figure*}

We now present our experiments, starting with a discussion of the datasets we use for training, followed by an illustration of qualitative fitting and relighting results, an ablation study to validate our design choices, and finally comparisons to existing methods.

\subsection{Datasets}
A key component of our method is that we can train on various existing multi-view face datasets despite different camera and lighting configurations.  For all our results, we use four datasets, as described below.

\begin{description}[
    leftmargin=0pt, 
    labelindent=0pt, 
    itemsep=0pt,
    listparindent=\parindent 
]
    \item[D1 - 3DPR~\cite{prao20253dpr}.]  This is the one dataset with OLAT illumination, which we use to train Stage 2 of our pipeline.  It consists of 40 cameras (we use 25 frontal cameras) and we processed the neutral expression for 127 subjects (116 for training and 11 for testing). The data contains 331 point lights, as well as fully-lit frames that we use to train Stage 1. 
    \item[D2 - Ava-256~\cite{martinez2024codec}.]  Consisting of 80 cameras (we use 55 frontal cameras), lit from 360 degrees. We processed 240 subjects and use one neutral frame for each subject in Stage 1 training.
    \item[D3 - SDFM~\cite{chandran2020semantic}.]  Consisting of 8 cameras arranged as 4 stereo pairs (we omit the cross-polarized cameras), lit from 4 frontal flashes.  We processed 151 subjects and use one neutral frame for each subject in Stage 1 training.
    \item[D4 - Nersemble~\cite{kirschstein2023nersemble}.] Consisting of 16 cameras, lit from 8 frontal flashes but has a light background that reflects light from behind. We processed 411 subjects and use one neutral frame from each subject in Stage 1 training.
\end{description}
Although some datasets come with tracked geometry, we run the VHAP face tracker~\cite{qian2024vhap, qian2024gaussianavatars} for all the datasets to obtain meshes in the same topology, in the same canonical space. We then crop each frame based on the mesh projection in the image plane to $1024 \times 1024$ resolution. We also compute a mask based on the mesh and matting~\cite{lin2021real, yao2024matte} to mask out the regions below the neck. Datasets D1, D3 and D4 provide color calibration but D2 does not. Therefore, we do a warmup run (2000 iterations) of the Stage 1 model without the lighting code $\mathbf{z}_l$ but instead optimize a $3 \times 3$ color matrix for D2, which we use afterwards to color calibrate the images in D2 and train again with the lighting code learning enabled.

\subsection{Qualitative Results}
Once trained, we can fit our model to unseen subjects to build 3D Gaussian head representations, and then relight those under any environment lighting.  Several results are shown in~\figref{fig:hero}.  The first 5 rows show training subjects from the flat-lit datasets (D2, D3 and D4), where we perform fitting on all input views.  The last 2 rows show fitting to single portrait images in the wild.   In all cases, the reconstructed 3DGS head avatar has good 3D consistency under novel view rendering, and can be relit in any outdoor or indoor environment.  Please see~\figref{fig:teaser} for additional results.  These fitting and relighting results show that our method generalizes to any identity, which is possible due to our two-stage pipeline designed to train across a variety of existing datasets, without the need for a large corpus of OLAT data.

\begin{figure}[t]
  \centering
  \includegraphics[width=\linewidth]{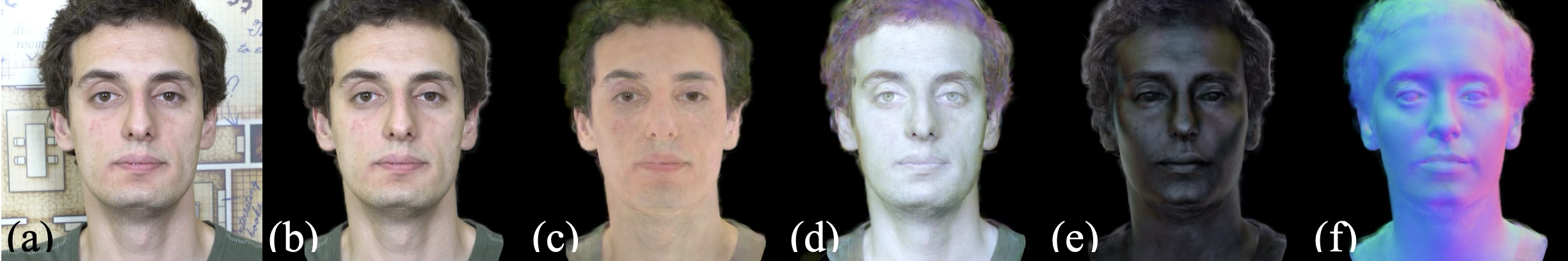}
   \caption{For a single image of an unseen subject in-the-wild  (a), our method predicts a Gaussian avatar (b) with an intrinsic decomposition into diffuse albedo (c), diffuse shading (d), specular component (e), and specular normals (f).}
   \label{fig:decomp}
\end{figure}

We also show the ability of our network to separate albedo from reflectance parameters for an unseen subject in~\figref{fig:decomp}, illustrating the learned intrinsic decomposition of an in-the-wild image with unknown lighting.

\subsection{Ablation}
\begin{figure}[t]
  \centering
  \includegraphics[width=\linewidth]{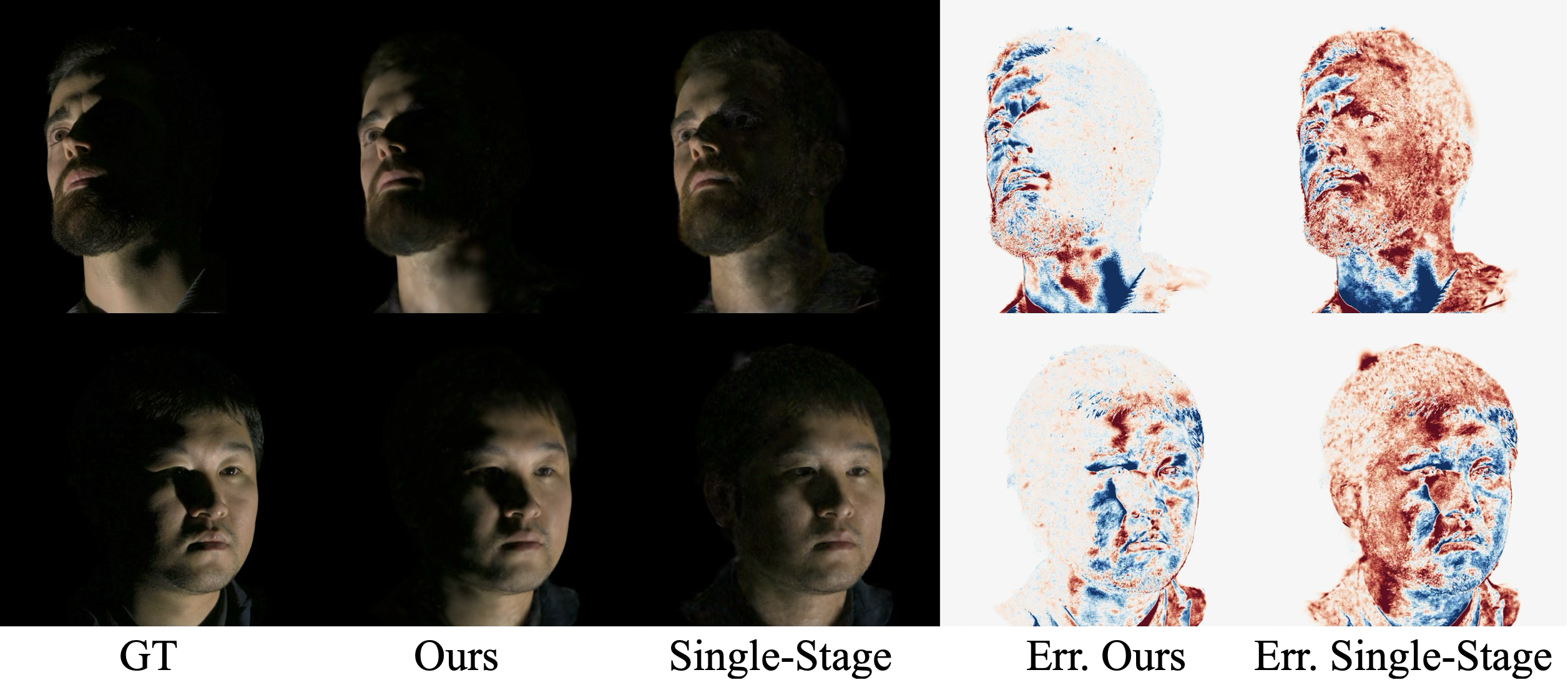}
   \caption{Ablation of the two-stage pipeline on the test subjects in D1 under OLAT lighting. The single-stage pipeline especially struggles in the shadowed areas. The render errors are in the range of $-0.1$~\includegraphics[width=1cm]{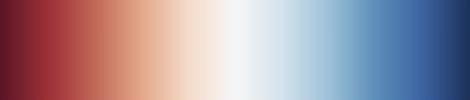}~0.1. Images best viewed zoomed-in.}
   \label{fig:one-stage-mpi}
\end{figure}

\begin{figure}[t]
  \centering
  \includegraphics[width=\linewidth]{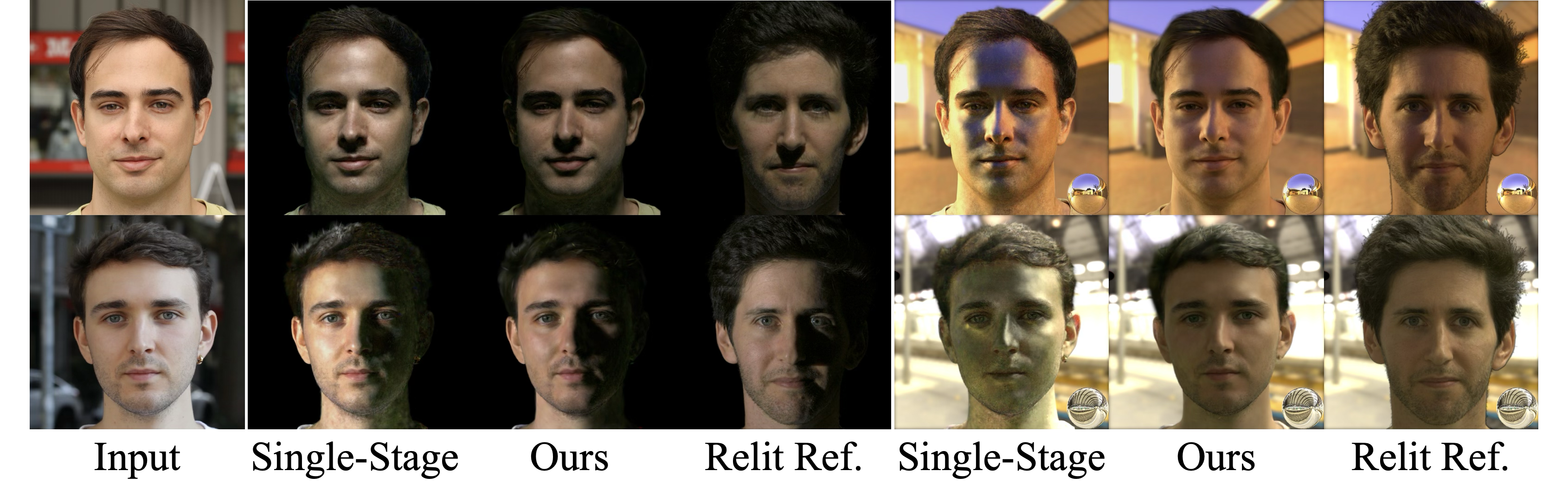}
   \caption{Ablation of the proposed two-stage pipeline \vs the single-stage pipeline on in-the-wild images. The single-stage pipeline struggles with preserving hard shadows which leads to severe artifacts in the environment relighting case.}
   \label{fig:one-stage-wild}
\end{figure}

\paragraph{Two-Stage \vs Single-Stage.}
A straight-forward way to improve over RGCA for multi-identity relighting is to directly add an identity code and keep it single-stage. More specifically, this single-stage model shares a similar structure as our Stage 1 model but it replaces $\mathcal{D}_c$ with two decoders  $\mathcal{D}'_{ci}$ and $\mathcal{D}'_{cv}$ that directly predict relightable 3D Gaussian parameters:
\begin{align}
\{\boldsymbol{\rho}_k, \mathbf{d}_k, \sigma_k\}_{k=1}^M &= \mathcal{D}'_{ci}(\mathbf{z}_\text{id})\\
\{v_k, \delta\mathbf{n}_k\}_{k=1}^M &= \mathcal{D}'_{cv}(\mathbf{z}_\text{id}, \boldsymbol{\omega}_o)
\end{align}
Note that this network design allows only training with OLAT datasets (\ie, D1), and it is closer to what is done in other related work~\cite{yang2024vrmm, li2024uravatar} as they have larger OLAT datasets. We train this single-stage model with the same train-test split. For each test subject, we fit the trained model on a fully-lit frame, and evaluate the relighting performance on the ground truth OLAT frames. Quantitative results are reported in~\tabref{tab:one-stage}, using image metrics including PSNR, RMSE, SSIM~\cite{wang2004image} and LPIPS~\cite{zhang2018unreasonable}. We also show some examples with the corresponding error maps in~\figref{fig:one-stage-mpi}. We notice that the single-stage model struggles in keeping hard shadows. We further demonstrate in~\figref{fig:one-stage-wild} that when the lighting is unknown and must be optimized, the single-stage model fails entirely in keeping the learned relighting prior and produces obvious relit artifacts.

\begin{table}
	\centering
    \caption{Ablation of using two-stage \vs single-stage design with image metrics on the test subjects in D1.}\label{tab:one-stage}
    \vspace*{-2mm}
	\footnotesize
	\begin{tabular}{l|cccc}
		\toprule
		& PSNR $\uparrow$ & RMSE $\downarrow$ & SSIM $\uparrow$ & LPIPS $\downarrow$ \\
		\midrule
		Single-Stage &  25.49 &  0.1092 &  0.76 & 0.2732 \\
	    Two-Stage (Ours) & \textbf{30.06} &  \textbf{0.0655} &  \textbf{0.87} &  \textbf{0.2358} \\
		\bottomrule
	\end{tabular}
\end{table}

\begin{figure}[t]
  \centering
  \includegraphics[width=\linewidth]{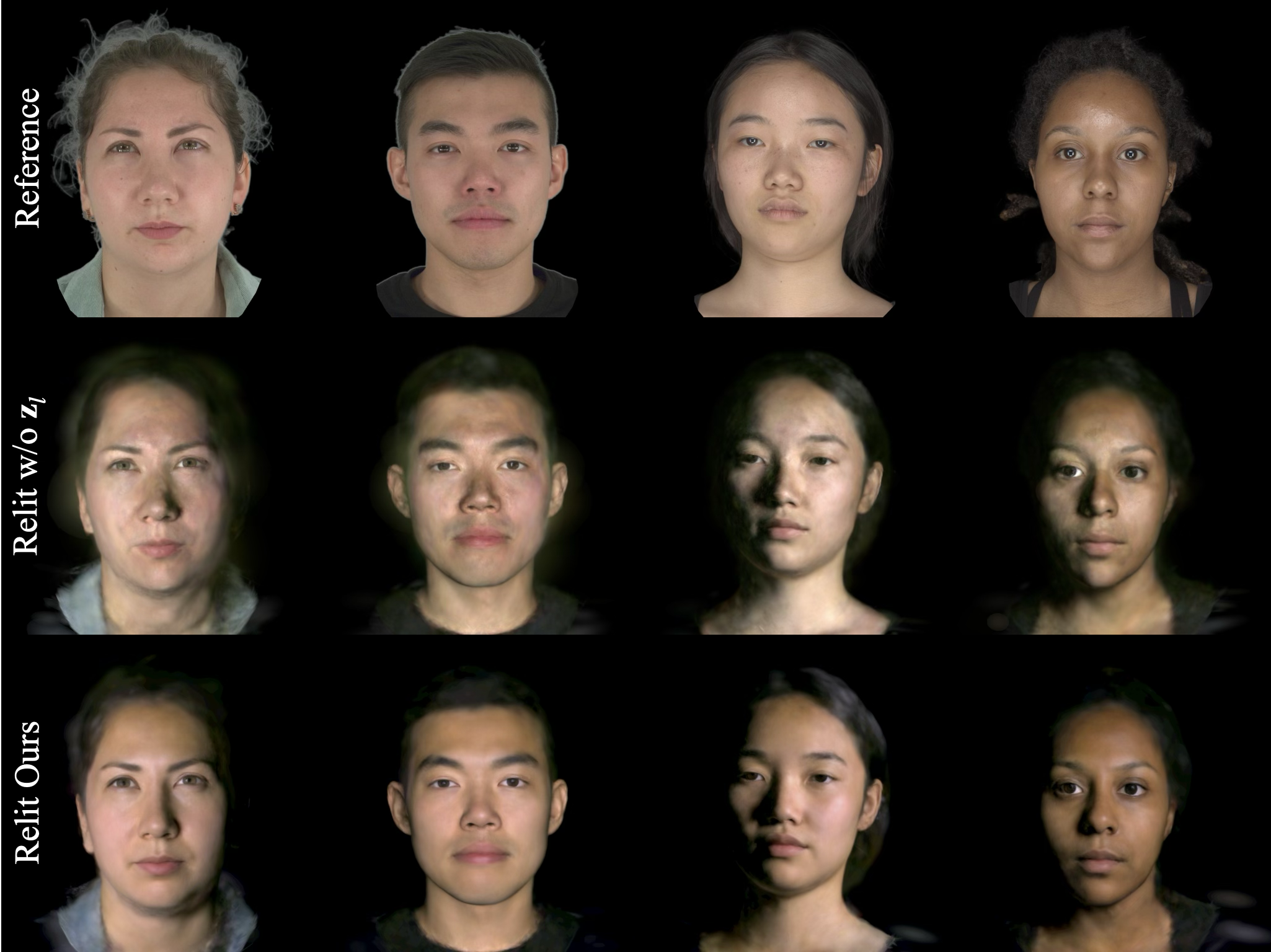}
   \caption{Ablation of the dataset-specific lighting code. Without $\mathbf{z}_l$, the lighting differences in the datasets are entangled in the representation, leading to blotchy artifacts during re-lighting.}
   \label{fig:setup-code}
\end{figure}

\paragraph{Lighting Code $\mathbf{z}_l$.}
We now ablate the effect of incorporating the dataset-specific lighting code. The ablated model modifies $\mathcal{D}_c$ to take only $\mathbf{z}_\text{id}$ as input. After training, we test the model by passing the full-on Gaussian textures of subjects from Stage 1 (who lack OLAT data, \ie, from D2, D3, or D4) through Stage 2 to get their relit version. Note that relit ground truth is not available for these subjects. We therefore only show qualitative point light relit examples. Without $\mathbf{z}_l$, the dataset-specific lighting impacts are entangled with the subjects' appearance. Because the relighting network is trained only on D1, passing subjects directly from D2, D3 or D4 results in blotchy artifacts in relit renders, as shown in~\figref{fig:setup-code}. In contrast, with our model, we can align all datasets to D1 by swapping the lighting code before passing them through the relighting network, resulting in smooth and photo-realistic relit renders.

\subsection{Comparisons}
\begin{figure}[t]
  \centering
  \includegraphics[width=\linewidth]{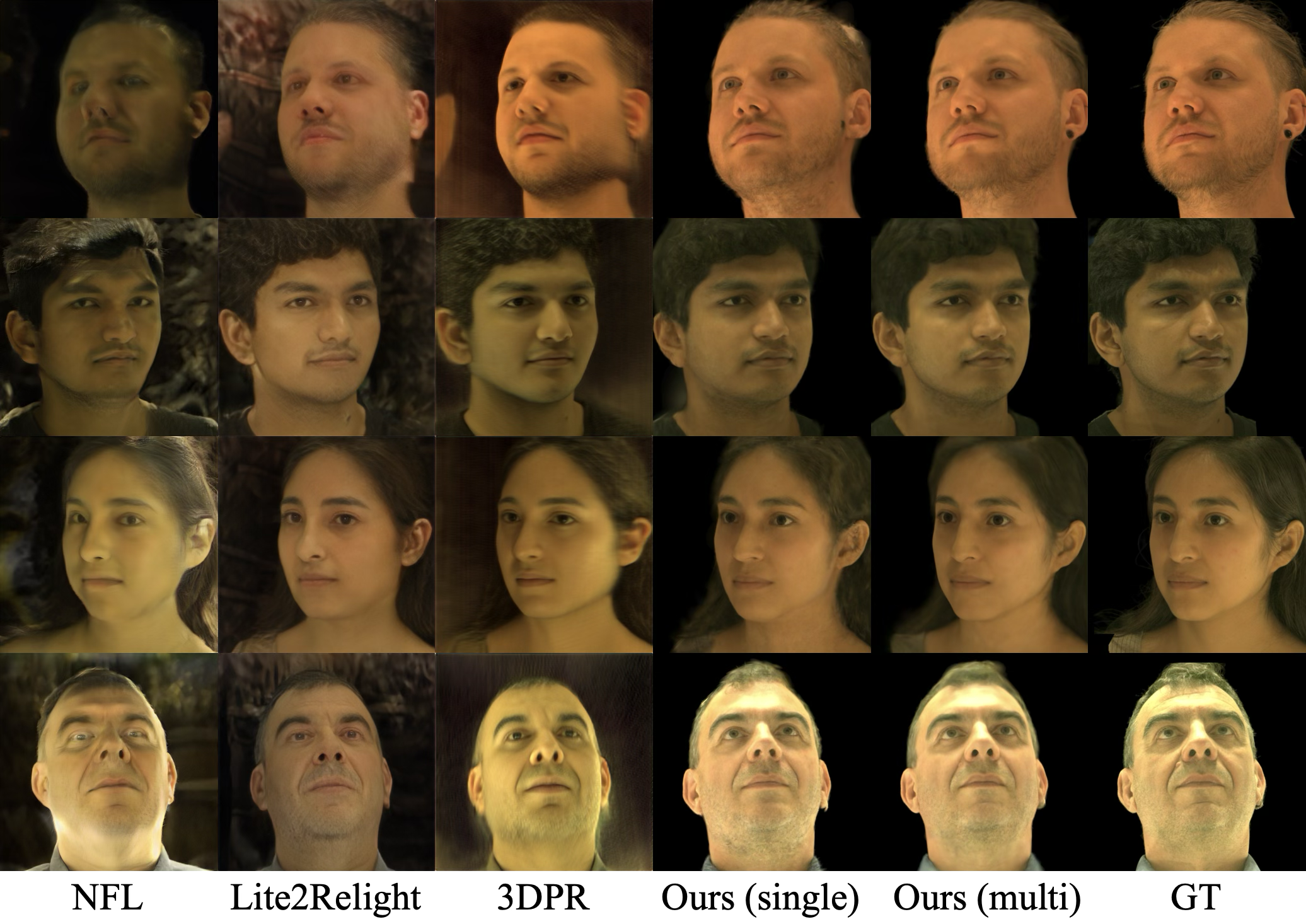}
   \caption{Examples from the comparison against 3D GAN-based relighting methods on test subjects in D1: Note how our proposed method matches the illumination of the ground truth most closely and preserves the most details.}
   \label{fig:3dpr-test}
\end{figure}

We compare our method with generalized relighting methods that are based on 3D-aware GANs, as well as 2D diffusion-based methods. Unfortunately, the method closest to ours (\ie, URAvatar~\cite{li2024uravatar}) has no code available.
\paragraph{3D GAN-Based Methods.} We first compare with 3D GAN-based relighting methods: NeRFFaceLighting (NFL)~\cite{jiang2023nerffacelighting}, Lite2Relight~\cite{rao2024lite2relight}, and a very recent state-of-the-art 3DPR~\cite{prao20253dpr}. We first report quantitative image metrics in~\tabref{tab:comp}, using the same test set as 3DPR and reuse their numbers from the original paper. Note that since some of the baselines cannot handle high-frequency relighting, we evaluate the relighting performance of the methods on the same set of low resolution ($10 \times 20$) environment maps used in 3DPR. The ground truth relit results are generated with image-based relighting~\cite{debevec2000acquiring}.  We show some image examples in~\figref{fig:3dpr-test}. We also note two shared limitations of these 3D GAN-based methods: (1) The heads are distorted for non-frontal poses as these methods are trained on 2D portrait collections without enforcing explicit multi-view consistency. (2) They cannot fit to multi-view images. In contrast, we can fit to both multi-view and a single image of a person and are fully view consistent. Please refer to the supplementary material for novel view synthesis comparison.

\begin{table}
	\centering
    \caption{Comparison against 3D GAN-based relighting methods on the test subjects in D1 with environment-map relighting. }\label{tab:comp}
	\footnotesize
    \vspace*{-2mm}
	\begin{tabular}{l|cccc}
		\toprule
		& PSNR $\uparrow$ & RMSE $\downarrow$ & SSIM $\uparrow$ & LPIPS $\downarrow$ \\
		\midrule
		NFL~\cite{jiang2023nerffacelighting} &  16.97 &  0.2926 &  0.77 & 0.2385 \\
		Lite2Relight~\cite{rao2024lite2relight} & 16.72 & 0.2619 &  0.79 & 0.2506 \\
		3DPR~\cite{prao20253dpr} & 21.02 & 0.1801 & 0.83 & 0.1996 \\
        Ours (single image) & 26.57 & 0.0996 & 0.86 & 0.1671 \\
	    Ours (multi-view) & \textbf{29.07} &  \textbf{0.0746} &  \textbf{0.91} &  \textbf{0.1649} \\
		\bottomrule
	\end{tabular}
\end{table}

\begin{figure}[t]
  \centering
  \includegraphics[width=\linewidth]{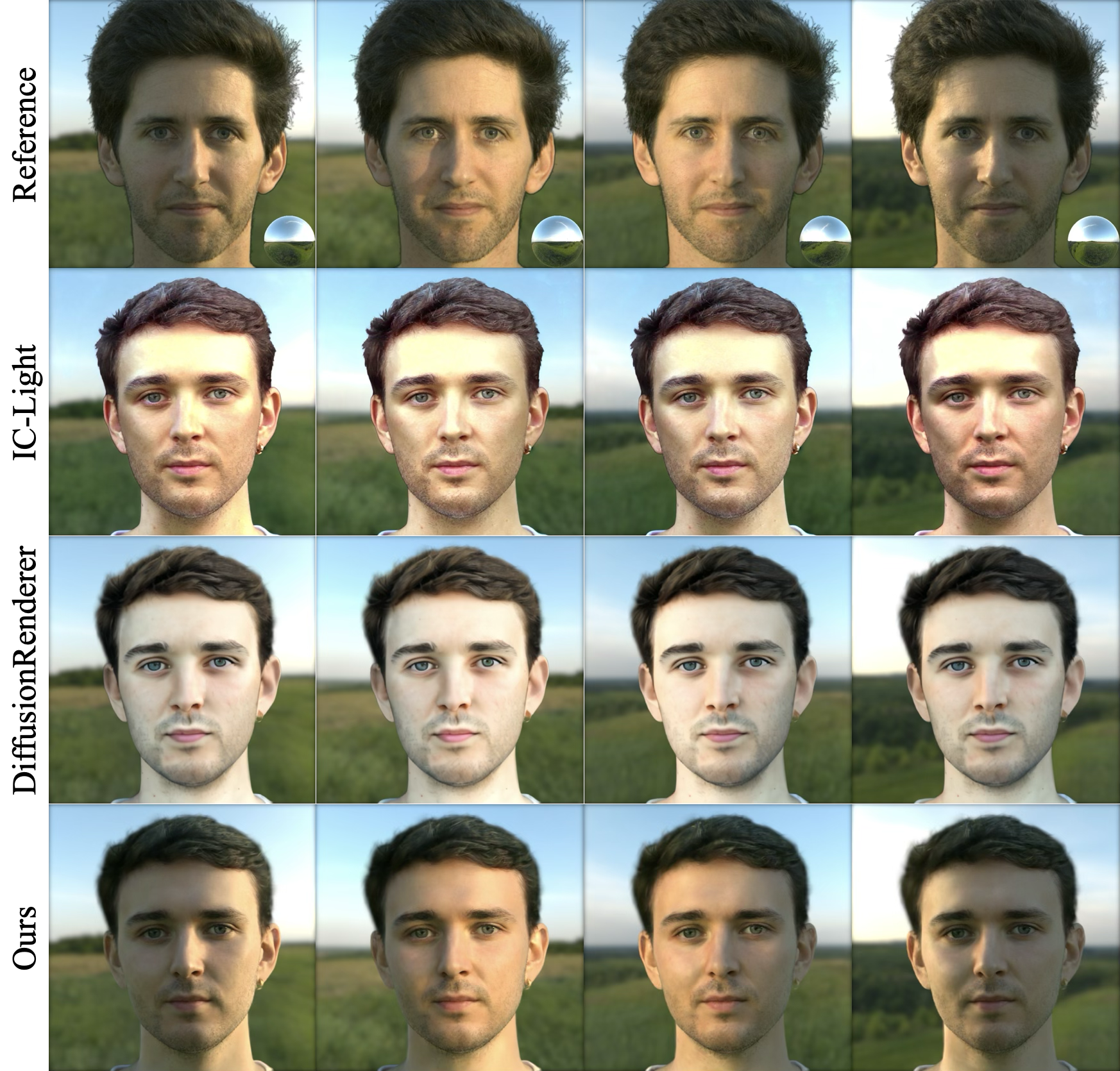}
   \caption{Comparison with 2D diffusion-based methods on relighting an avatar with a rotating environment map. IC-Light relights the image based on the background and fails to match the main light. DiffusionRenderer estimates an overly-diffuse skin material.}
   \label{fig:iclight}
\end{figure}

\paragraph{2D Diffusion-Based Methods.}
We also compare against some recent diffusion-based techniques, IC-Light~\cite{zhang2025scaling} and DiffusionRenderer~\cite{DiffusionRenderer}. Note that these are purely 2D relighting methods, which means they cannot generate novel views of a subject. This is already a limitation compared to ours. IC-Light (background-conditioned model) only learns to relight the foreground such that it matches with the provided background. It lacks a way for fine-grained lighting control such as using an HDRI map. In~\figref{fig:iclight}, we show frames from a sequence relit by a rotating environment map. IC-Light fails to capture the main light source and the relit results are incoherent across frames, also exhibiting an unnatural metallic sheen on the skin. DiffusionRenderer tackles relighting with an inverse rendering approach, first estimating the intrinsic material properties of the scene. However, the estimated skin reflectance tends to be overly diffuse, as shown in \figref{fig:iclight} and it also fails to render realistic shadows.

\section{Conclusion}
\label{sec:conclusion}

We present \emph{RelightAnyone}, a new model for 3DGS head avatar reconstruction and relighting.  Unlike previous methods, our approach is based on the unification of multiple existing multi-view face datasets.  Our novel two-stage design allows us to train a flat-lit Gaussian reconstruction stage on datasets without OLAT illumination, followed by a mapping network that learns to infer physically-based relightable parameters for flat-lit avatars, trained on substantially less time-multiplexed OLAT data.  This is possible due to our strategy for self-supervised lighting alignment across datasets. However, our model still has limitations. First, it struggles with hair reconstruction and relighting, primarily due to inaccuracies in tracked hair geometry and the difficulty of establishing reliable UV correspondences for hair strands. Future work could consider a separate model for hair and face, as in~\cite{kim2025haircup}. Second, our model is currently trained only on the neutral expression. Extending it to dynamic performances and capturing expression-dependent appearance would require a larger dataset with diverse expressions captured under both OLAT and fully-lit conditions. 
Nevertheless, we present a powerful model that can be fit to unseen subjects in unseen environments, from as little as a single image in-the-wild, with superior performance over previous state-of-the-art methods.
{
    \small
    \bibliographystyle{ieeenat_fullname}
    \bibliography{main}
}

\clearpage
\setcounter{page}{1}
\maketitlesupplementary

\section{Implementation Details} 
\paragraph{Network Details.} 
Our 2D convolutional decoders, \ie, $\mathcal{D}_g$, $\mathcal{D}_c$, $\mathcal{D}_{ci}$ and $\mathcal{D}_{cv}$, have a nearly identical architecture, differing only in their specific input/output layers and skip connections. The input vector is first linearly mapped and reshaped into an initial feature map $\mathbf{z}' \in \mathbb{R}^{256 \times 8 \times 8}$ (channels $\times$ height $\times$ width). Then, at each layer, it is progressively upsampled by a factor of two until it reaches the final $1024 \times 1024$ resolution. All intermediate layers are followed by LeakyReLU activations. $\mathcal{E}$ is a mirrored version of $\mathcal{D}_{ci}$ and $\mathcal{D}_{cv}$. We apply specific activation functions to the final output: a softplus function for the Gaussian scales $\mathbf{s}_k$, a sigmoid function for opacity $o_k$ and specular visibility $v_k$, and an exponential function for the roughness $\sigma_k$. Gaussian colors are clamped to be non-negative before splatting. 

\paragraph{Training Details.} 
We set the loss balancing weights as follows: $\lambda_\text{l1} = 10$, $\lambda_\text{ssim} = 0.2$, $\lambda_\text{geo} = 0.4$, $\lambda_s = 0.01$, $\lambda_{c_-} = 0.01$, $\lambda_\text{mono} = 0.01$, $\lambda_\text{id} = 0.01$, and $\lambda_\text{lr} = 1$. Several weights are linearly annealed: $\lambda_t$ is initialized at 1 and decreased to 0.001 by iteration 20000; $\lambda_n$ is initialized at 1 and decreased to 0 by iteration 5000;  $\lambda_\rho$ is initialized as 10 and decreased to 0.01 by iteration 10000. We use the Adam optimizer with a learning rate of $1e^{-3}$ to for Stage 1, and $5e^{-4}$ for Stage 2 and model fitting. A batch size of 16 is used for both stages. Both Stage 1 and Stage 2 models are trained for one day on 4 Quadro RTX 6000/8000 GPUs. The model fitting process, including both the inversion and finetuning steps, typically converges within 3000 iterations, taking approximately 30 minutes on a single GPU.

\section{Additional Experiments}
\begin{figure}[t]
  \centering
  \includegraphics[width=\linewidth]{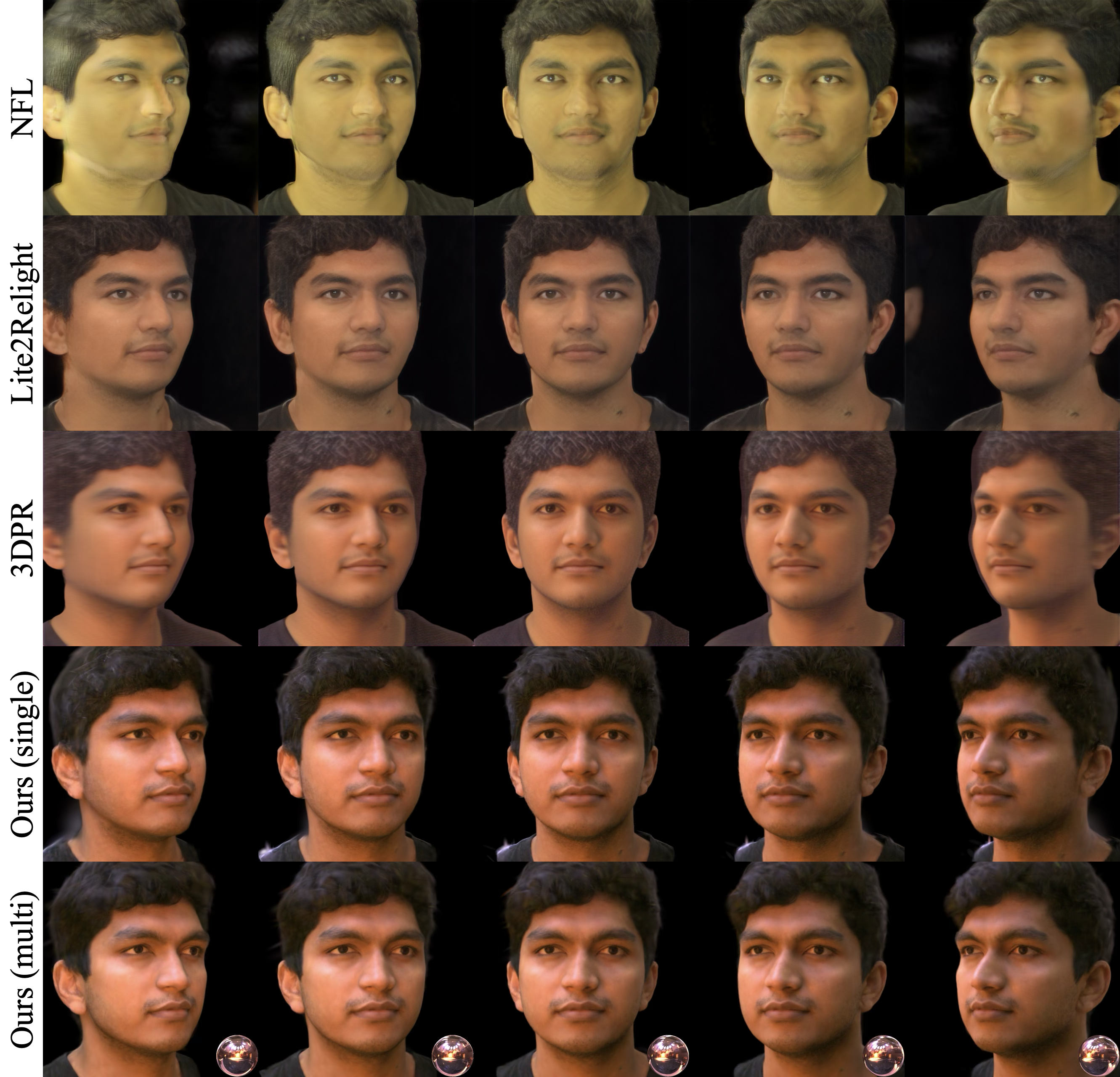}
   \caption{Novel view comparison with 3D GAN-based relighting methods: Since the baseline methods are trained only on 2D portrait collections, they struggle with view-consistency at the side poses. Furthermore, since the baselines can only take a single image input, we also show our method fitted to only a single image which slightly degrades the quality in side poses.}
   \label{fig:3dpr-mv}
\end{figure}

\paragraph{Novel View Comparison.}
\figref{fig:3dpr-mv} compares our novel view synthesis with 3D GAN-based methods: NeRFFaceLighting (NFL)~\cite{jiang2023nerffacelighting}, Lite2Relight~\cite{rao2024lite2relight} and 3DPR~\cite{prao20253dpr}, which are trained only on 2D portrait collections, often produce distorted or ``stretched'' results for side poses. Moreover, they cannot be applied trivially to multi-view inputs of the same subject, as they typically encode each view into a different latent vector. In contrast, our method learns an explicit volumetric representation directly from multi-view data, resulting in better view-consistency. We note that when fitted to a single image, our method degrades only slightly in side poses, particularly in reconstructing the ears and the facial silhouette. Our results also better preserve the identity (see~\figref{fig:3dpr-test} for a real photo of this subject).

\begin{figure}[t]
  \centering
  \includegraphics[width=\linewidth]{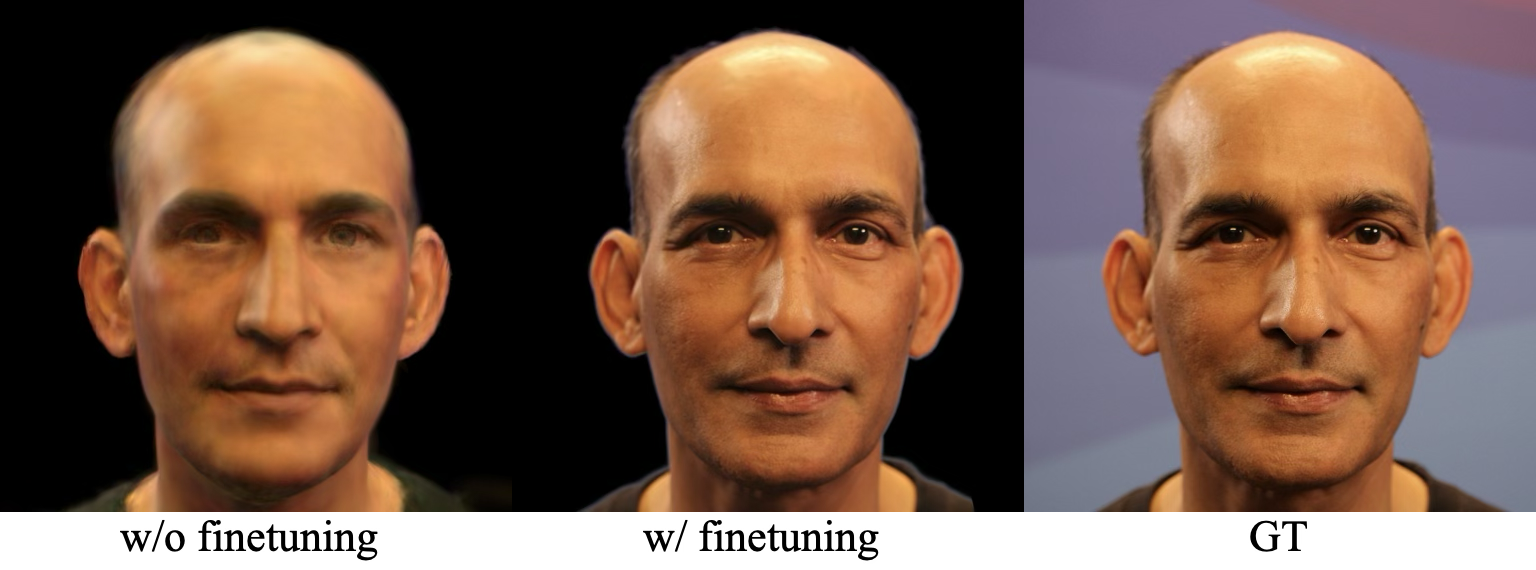}
   \caption{Effect of finetuning. While optimizing only the identity code and lighting produces a rough likeness, finetuning our model recovers high-frequency, person-specific details that better match the ground truth input.}
   \label{fig:finetune}
\end{figure}

\paragraph{Effect of Finetuning.}
\figref{fig:finetune} demonstrates the effect of finetuning on a single in-with-wild input image. Without finetuning, \ie, optimizing only the identity code and the scene lighting (see ``w/o finetuning'' column), the rendered image captures only a rough likeness of the subject with low-frequency appearance. By finetuning the Stage 1 model, we can capture more person-specific details, resulting in a rendered image more closely matches the ground truth.

\begin{figure}[t]
  \centering
  \includegraphics[width=\linewidth]{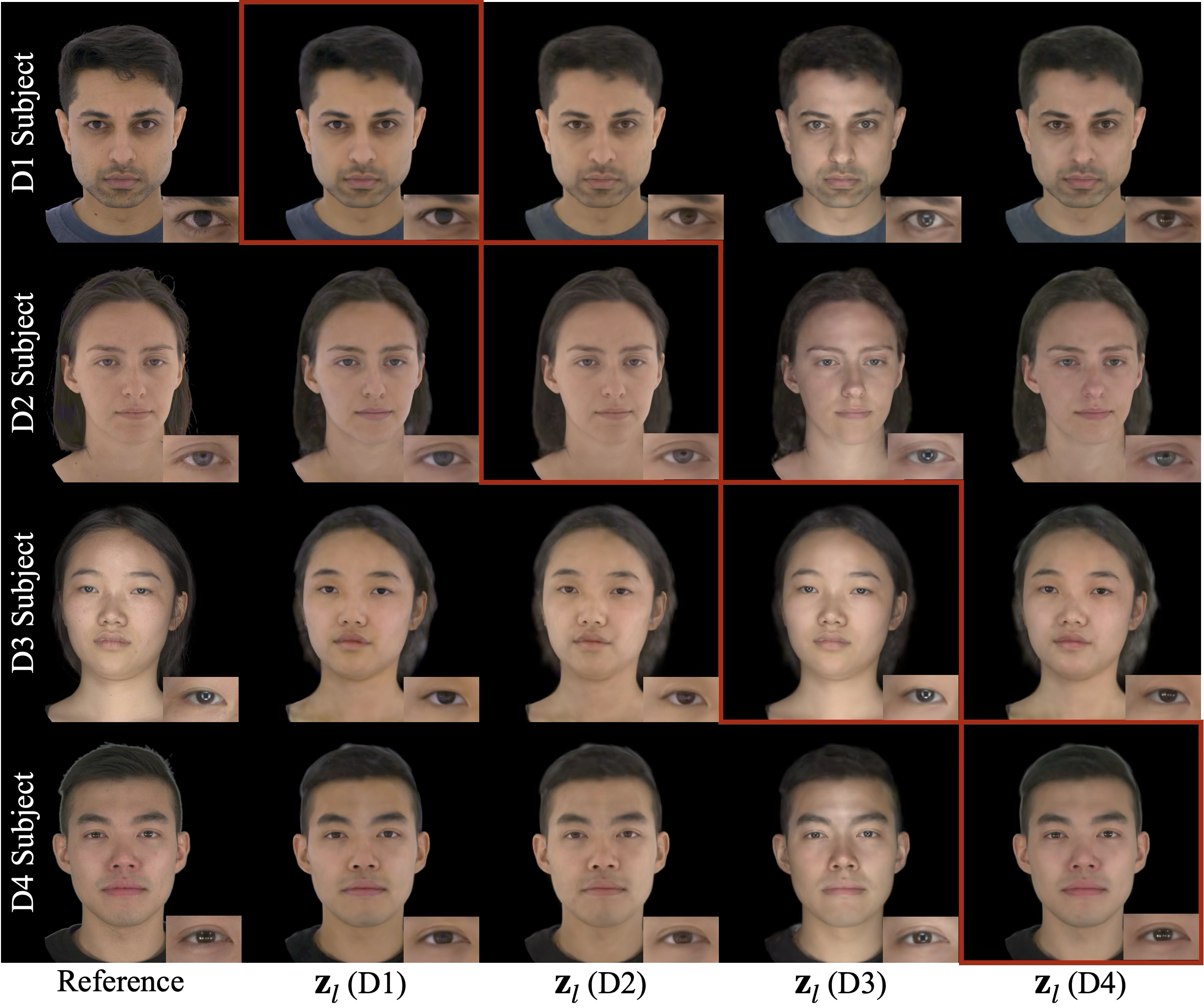}
   \caption{Self-supervised lighting alignment. Our model learns a distinct lighting code $\mathbf{z}_l$ for each dataset. Each row shows a subject rendered with the lighting codes from different datasets (D1-D4). Red boxes indicate the subject's original dataset. Note how our model captures dataset-specific lighting variations, especially in the specular reflections on the nose and in the eyes. Image best viewed zoomed-in.}
   \label{fig:light-align}
\end{figure}

\paragraph{Lighting Alignment.}
Our model enables self-supervised lighting alignment by introducing a dataset-specific lighting code $\mathbf{z}_l$. As shown in~\figref{fig:light-align}, each row corresponds to a subject from a different dataset (D1, D2, D3 and D4, from top to bottom). The first column shows the ground truth images for reference, and the subsequent columns show fully-lit renders generated using the lighting codes from each of the four datasets. Red boxes indicate the original lighting condition for each subject. Although all datasets are generally evenly-lit, our $\mathbf{z}_l$ code successfully learns their subtle, distinct lighting distributions. For example, subjects in dataset D3 are lit from four frontal flashes, resulting in stronger specular highlights in the central part of the face. Our model correctly captures this specific effect when applying the D3 lighting code. Similarly, while dataset D4 is also front-lit, it has a light background that reflects light from behind, making it closer to D1 and D2 (which are lit from 360 degrees). Even so, our model is able to capture the subtle differences in specular reflections on the nose and in the eyes.

\begin{figure}[t]
  \centering
  \includegraphics[width=\linewidth]{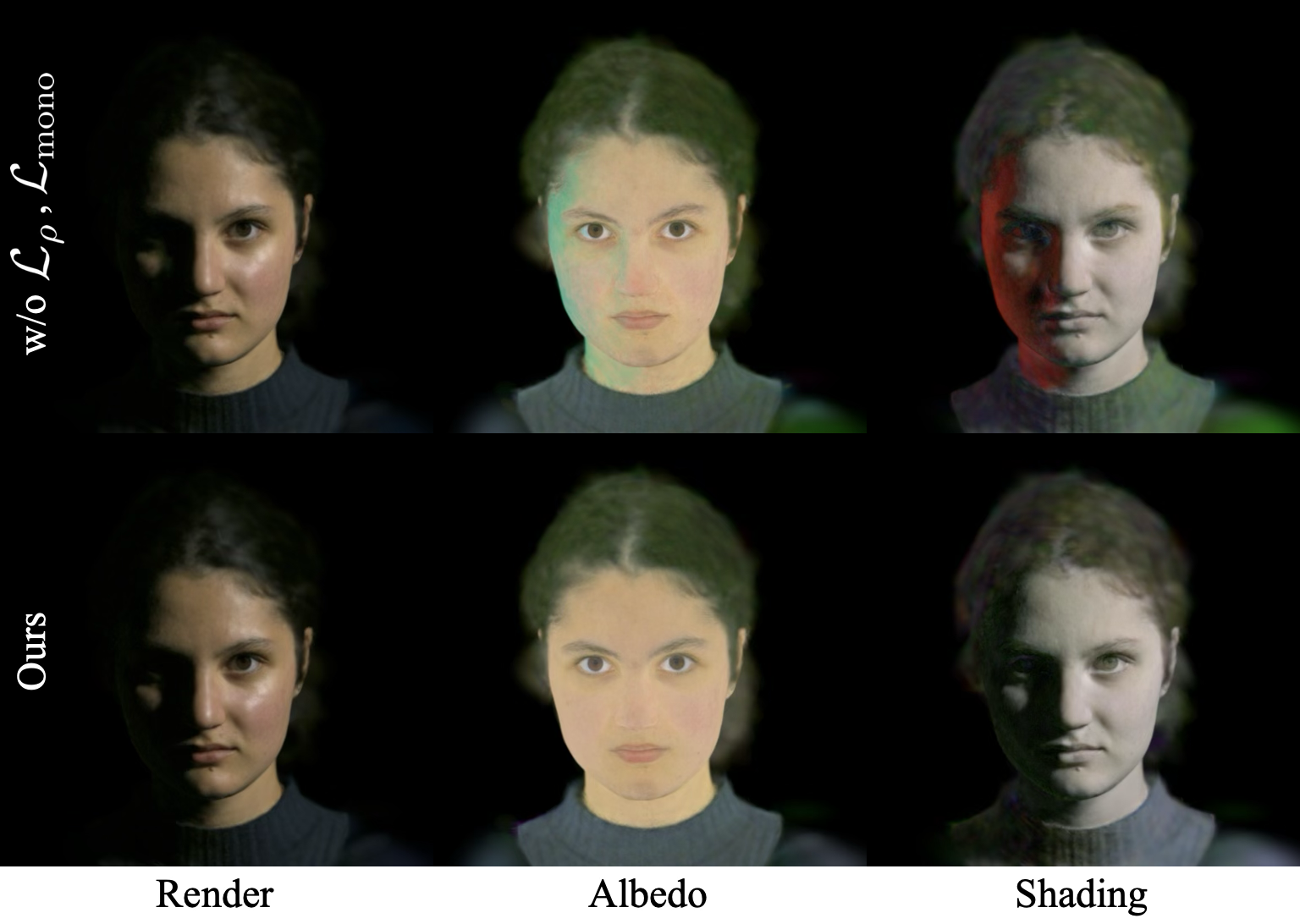}
   \caption{Effect of $\mathcal{L}_\rho$ and $\mathcal{L}_\text{mono}$ on intrinsic decomposition. Without our regularization (top row), the model produces a plausible render but fails to properly disentangle albedo and shading. Our full model (bottom row) achieves a clean and physically meaningful intrinsic decomposition.}
   \label{fig:albedo-shading}
\end{figure}

\paragraph{Effect of $\mathcal{L}_\rho$ and $\mathcal{L}_\text{mono}$.}
\figref{fig:albedo-shading} demonstrates the effect of the regularization terms $\mathcal{L}_\rho$ and $\mathcal{L}_\text{mono}$, which we introduced to enforce a meaningful decomposition of the diffuse albedo and the diffuse shading. Without these regularization terms, the final render may look plausible, but the underlying albedo and diffuse shading components exhibit severe color artifacts. This occurs because the model gets stuck in a local minima, which it cannot escape as training proceeds. Our loss terms are designed to prevent this, guiding the optimization toward a correct decomposition.

\begin{figure}[t]
  \centering
  \includegraphics[width=\linewidth]{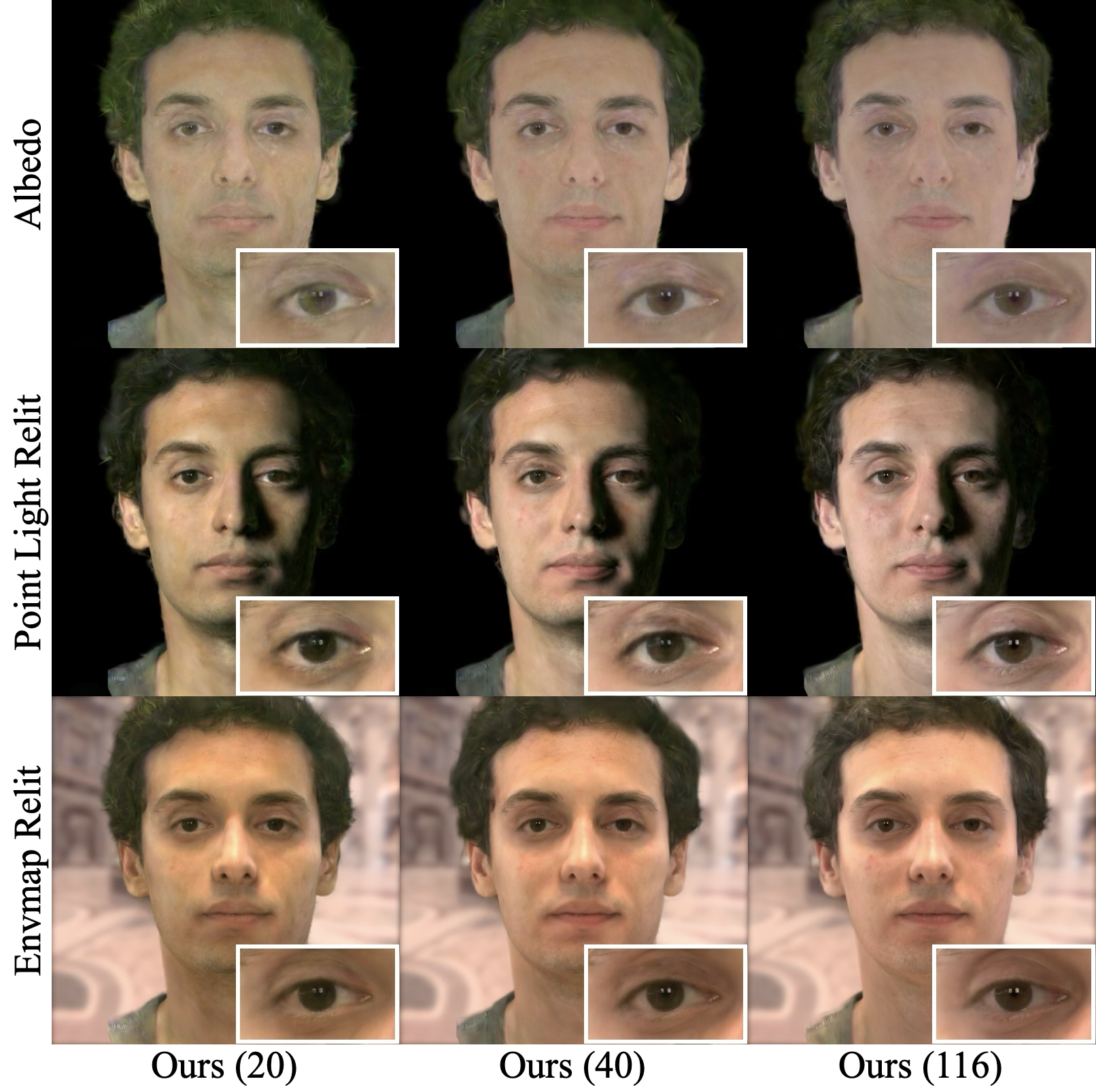}
   \caption{Effect of OLAT dataset size. The figure ablates the number of OLAT subjects used to train our Stage 2 network (20, 40, and 116 subjects, from left to right). While all models render the correct lighting distribution, training with more subjects produces a cleaner albedo and finer details which better preserves the subject's identity. Refer to~\figref{fig:decomp} for a real photo of this subject.}
   \label{fig:data-size-olat}
\end{figure}

\paragraph{Training with less OLAT Data.}
To evaluate the impact of the OLAT dataset size, we train our Stage 2 relighting network on subsets of 20, 40 and full 116 subjects in dataset D1. \figref{fig:data-size-olat} shows their fitting results on a single in-the-wild image. We note that all models successfully capture the correct lighting distribution, including similar shadows and highlights, which demonstrates the strong generalizability of our model, even when trained with minimal OLAT data. However, adding more OLAT data improves the results: the model trained on more subjects learns a cleaner albedo and better preserves identity (\eg, correct skin tone) and finer details.

\begin{figure}[t]
  \centering
  \includegraphics[width=\linewidth]{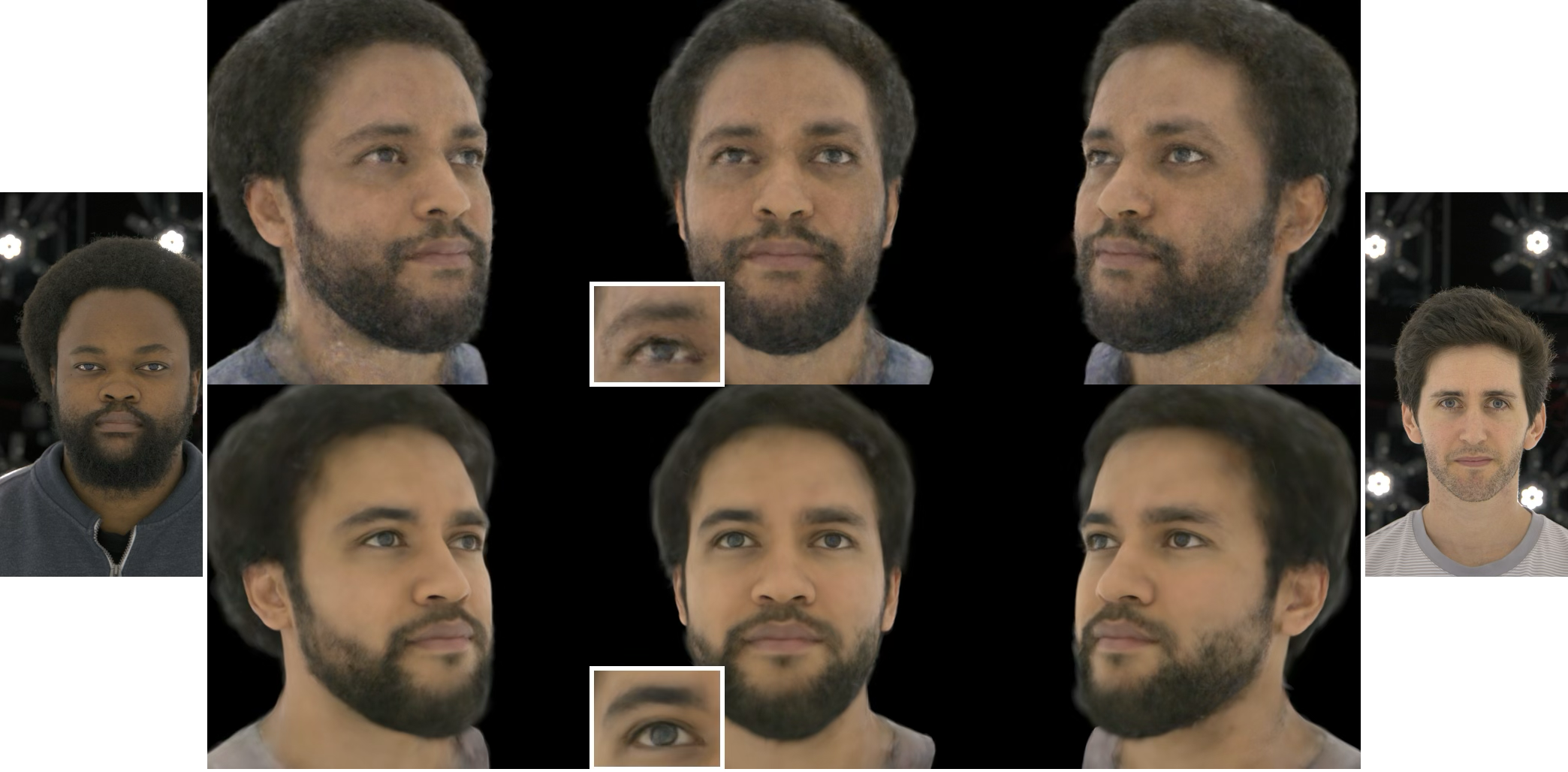}
   \caption{Novel view renders of an interpolated identity. Top row: Our Stage 1 model trained only on dataset D1 exhibits high-frequency artifacts. Bottom row: Our model trained on all four datasets yields a cleaner appearance. The references images for the two source identities used for interpolation are shown in the leftmost and rightmost columns.}
   \label{fig:interp-id}
\end{figure}

\paragraph{Training with less Full-On Data.}
Combining multiple existing flat-lit datasets improves the quality of the identity latent space and the learned multi-view prior. We demonstrate this in~\figref{fig:interp-id} by visualizing novel view renders of an interpolated identity. We compare a model trained only on dataset D1 (top row of~\figref{fig:interp-id}) to our full model trained on all four datasets (bottom row of~\figref{fig:interp-id}). We can see that training with the combined datasets produces a ``cleaner'' and more plausible new identity. In contrast, the ablated model (trained on D1 only) exhibits significant high-frequency artifacts, indicating a less robust latent space.

\begin{figure}[t]
  \centering
  \includegraphics[width=\linewidth]{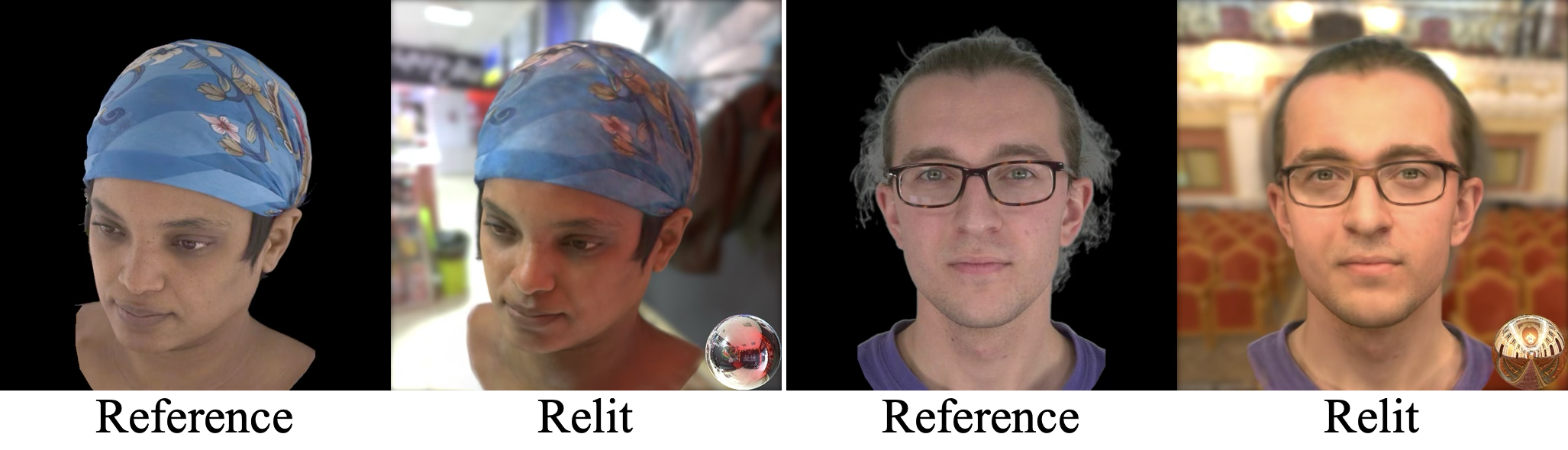}
   \caption{Failure case: accessories. Our model fails to infer relightable parameters for items like headscarf and glasses.}
   \label{fig:fail1}
\end{figure}

\paragraph{Failure Cases.}
Finally, we show some failure cases of our method. The first type of failure is associated with accessories, such as the headscarf and glasses shown in \figref{fig:fail1}. Because the OLAT dataset (\ie, D1) does not contain these accessories, our model cannot infer their relightable parameters. As a result, the patterns on the relit headscarf appear blurred, and the glasses lack specular reflections. We note that this is also a limitation of RGCA, as its appearance model is designed for the human head and does not work well on the diverse materials found in accessories. 

\begin{figure}[t]
  \centering
  \includegraphics[width=\linewidth]{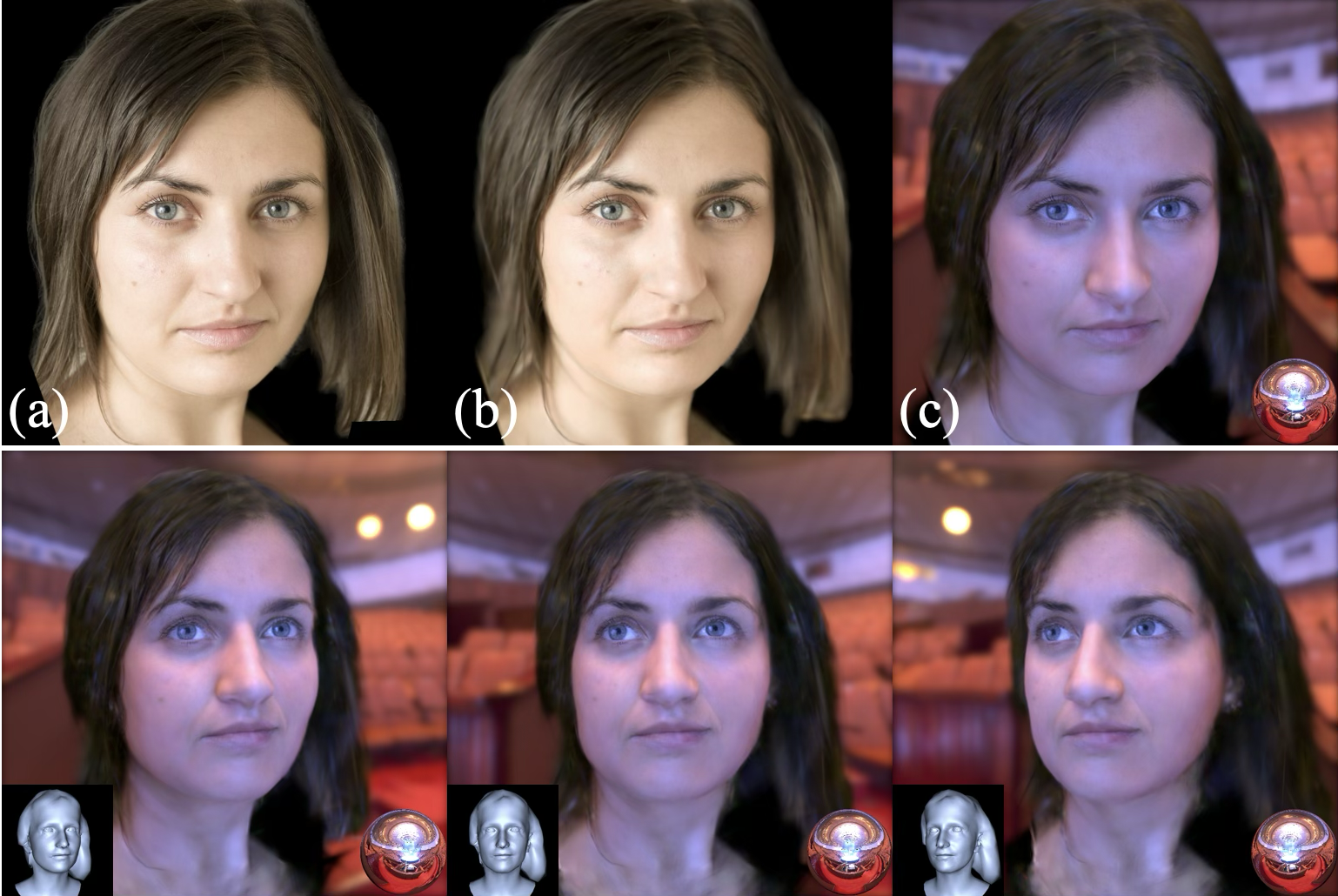}
   \caption{Failure case: open long hairstyle. Given an in-the-wild image of a subject with open long hair (a), the model fitting (b) and the original-view relit (c) appear plausible, but the hair appears as a texture-less cloud with color artifacts when rendered from novel views (bottom row). The corresponding tracked meshes are shown in the corner. Image best viewed zoomed-in.}
   \label{fig:fail2}
\end{figure}

Second, our model struggles with the reconstruction and relighting of some hairstyles. We show an example in~\figref{fig:fail2}, where our model can be fitted closely to a subject and relight them plausibly from the original camera view, but the hair appears as a texture-less cloud. This is especially visible when rendering novel views under new environment lighting, where some Gaussians also exhibit distracting color artifacts. There are several causes: first, although the VHAP~\cite{qian2024vhap} face tracker deforms the FLAME template to cover the hair, the results are sometimes poor for subjects with long hair (see inset). Second, these inaccurate tracking results lead to bad UV correspondences, making it difficult to learn a universal relightable prior for various hairstyles. Third, FLAME UV parameterization compresses the hair region into a small area on the UV map, allocating an insufficient number of Gaussians to represent the intricate structures. 

\section{Ethics}
All individuals portrayed in this paper provided informed consent for the use and publication of their images for research purposes.

\end{document}